\documentclass[10pt,journal,compsoc]{IEEEtran}

\usepackage{url}
\usepackage{microtype}
	
\usepackage{color, colortbl}
\ifCLASSOPTIONcompsoc
  \usepackage[nocompress]{cite}
\else
  \usepackage{cite}
\fi

\ifCLASSINFOpdf
\else
\fi

\usepackage{graphicx}
\usepackage{amsmath}
\usepackage{amssymb,amsfonts}
\usepackage{enumitem}
\usepackage{microtype}
\usepackage{bbm}
\usepackage{subfig}
\usepackage[
singlelinecheck=false 
]{caption}
\usepackage{booktabs}
\usepackage{multirow}
\usepackage[normalem]{ulem}
\usepackage[dvipsnames]{xcolor}
\usepackage[subtle]{savetrees}

\begin{document}
%
\title{Reconstruction guided Meta-learning for Few Shot Open Set Recognition}
%
%
%
%

\author{Sayak~Nag,
        Dripta~S.~Raychaudhuri,
        Sujoy~Paul,
        and~Amit~K.~Roy-Chowdhury,~\IEEEmembership{Fellow,~IEEE}}

\IEEEtitleabstractindextext{%
\begin{abstract}
In many applications, we are constrained to learn classifiers from very limited data (few-shot classification). The task becomes even more challenging if it is also required to identify samples from unknown categories (open-set classification). Learning a good abstraction for a class with very few samples is extremely difficult, especially under open-set settings. As a result, open-set recognition has received limited attention in the few-shot setting. However, it is a critical task in many applications like environmental monitoring, where the number of labeled examples for each class is limited. Existing few-shot open-set recognition (FSOSR) methods rely on thresholding schemes, with some considering uniform probability for open-class samples. However, this approach is often inaccurate, especially for fine-grained categorization, and makes them highly sensitive to the choice of a threshold. To address these concerns, we propose Reconstructing Exemplar-based Few-shot Open-set ClaSsifier (ReFOCS). By using a novel exemplar reconstruction-based meta-learning strategy ReFOCS streamlines FSOSR eliminating the need for a carefully tuned threshold by learning to be self-aware of the openness of a sample. The exemplars, act as class representatives and can be either provided in the training dataset or estimated in the feature domain. By testing on a wide variety of datasets, we show ReFOCS to outperform multiple state-of-the-art methods.
\end{abstract}

}

\maketitle

\IEEEdisplaynontitleabstractindextext

\IEEEpeerreviewmaketitle

\section{Introduction}
Deep neural networks have achieved excellent performance on a wide variety of visual tasks \cite{he2016deep,lin2017feature,long2015fully}. However, the majority of this success has been realized under the closed-set scenario, where it is assumed that all classes that appear during inference are present in the training set. In real-world applications, it is often difficult to obtain samples that exhaustively cover all possible semantic categories \cite{openlongtailrecognition}. This inherent open-ended nature of the visual world restricts the wide-scale applicability of deep models and machine learning models in general. Thus, it is more realistic to consider an \textit{open-set} scenario \cite{geng2020recent}, where the predictive model is expected to not only recognize samples from the seen classes, but also to recognize when it encounters an \textit{out-of-distribution} sample and reject it, rather than making a prediction for it.

\begin{figure}
    \centering
    \includegraphics[width=0.48\textwidth]{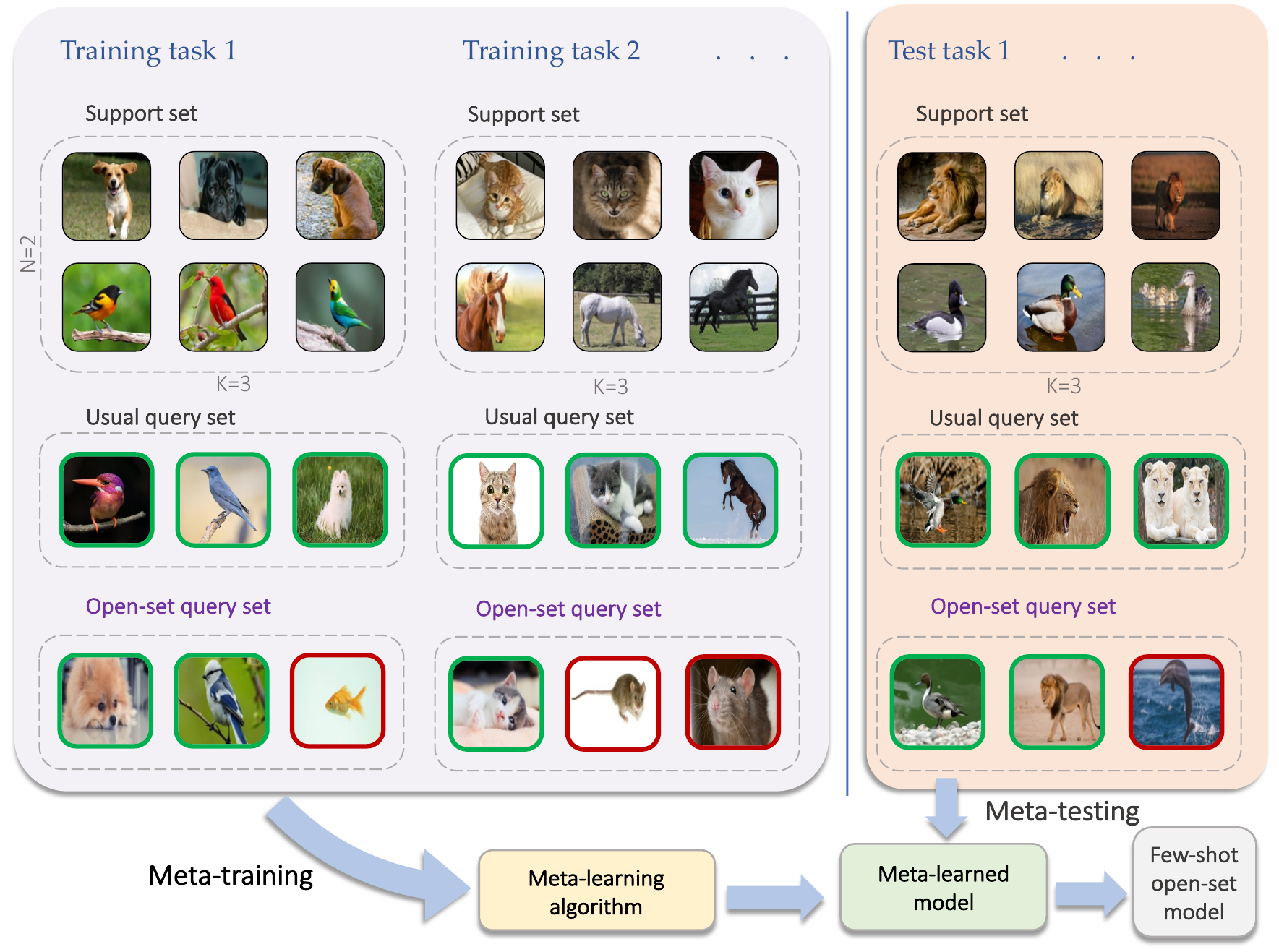}
    \caption{\textbf{Problem setup.} We formulate few-shot open-set recognition as a meta-learning problem where training proceeds in an episodic manner. In contrast to the usual setup, where the support and query sets share the same categories, we consider the more challenging open-set scenario where the query set can contain samples from classes not seen in the support (highlighted in red).}
    \label{fig:prob_setup}
\end{figure}

Notable approaches for open-set recognition involve adversarial training \cite{Lu_2017_ICCV} to reject adversarial samples which are too hard to classify, inducing self-awareness in CNNs \cite{hendrycks17baseline} to reject out-of-distribution samples, and using extreme value statistics to re-calibrate classification scores of samples from novel classes \cite{open_max}. All of these approaches require large amounts of labeled data per category for the seen classes. On the contrary, humans can easily grasp new concepts with very limited supervision and simultaneously perceive the occurrence of unforeseen abnormalities. Aiming to emulate this, we seek to perform open-set recognition in the \textit{few-shot learning} scenario, which has been largely ignored in the literature. A visual description of the problem setting is shown in Fig. \ref{fig:prob_setup}. 

\begin{figure}
    \centering
    \includegraphics[width=0.48\textwidth]{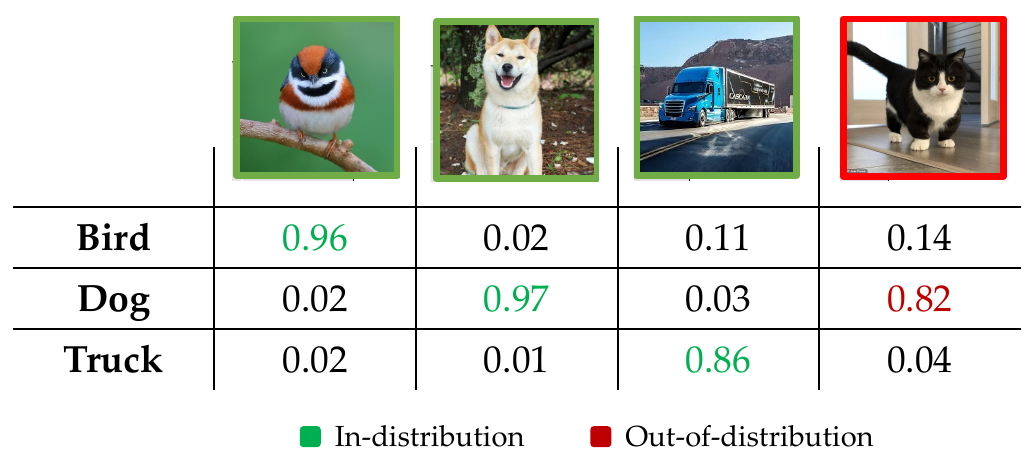}
    \caption{\textbf{Assuming uniform predictive distribution for outliers.} Out-of-distribution samples can be correlated to in-distribution classes by varying amounts. In this figure, we train a classifier for three in-distribution classes: dog, bird and truck (corresponding samples highlighted in green). The out-of-distribution sample - a cat - shares highly similar visual characteristics to a dog, as evidenced by the prediction. This suggests that desiring outliers to have a uniform predictive distribution, as suggested in \cite{peeler}, is often inaccurate.} 
    \label{fig:softmax_issues}
\end{figure}

The challenge in few-shot open-set recognition (FSOSR) stems from the limited availability of samples for in-distribution classes. This complicates learning a good abstraction of the provided categories to comprehensively distinguish between low-likelihood in-distribution samples and actual out-of-distribution samples. Following the success of meta-learning approaches \cite{maml,match_net,proto_net} in closed-set few-shot recognition, recent works \cite{peeler,snatcher} attempt FSOSR by building on top of the popular Prototypical Network framework \cite{proto_net}. In \cite{peeler}, the authors employ entropy maximization to enforce uniform predictive distribution of out-of-distribution samples to maximize model confusion for unseen categories. Out-of-distribution samples are identified by thresholding the maximum probability of the classification logits. In contrast, \cite{snatcher} leverages a transformer \cite{vaswani2017attention} and carefully crafted normalization techniques \cite{snatcher} to enforce improved representation consistency of in-distribution samples, enabling the rejection of out-of-distribution samples via simple distance thresholding in the feature space. Despite their effectiveness, these approaches rely heavily on carefully tuned threshold values. Furthermore, assuming near uniform predictive distribution for out-of-distribution samples does not generally hold in practice due to the sensitivity of deep CNNs to minor perturbations in the input \cite{goodfellow_adv} along with the tendency of the softmax operator to produce high confidence false predictions for out-of-distribution samples \cite{hendrycks17baseline}. This is further explained in Fig. \ref{fig:softmax_issues}.

\begin{figure}
    \centering
    \includegraphics[width=0.2\textwidth]{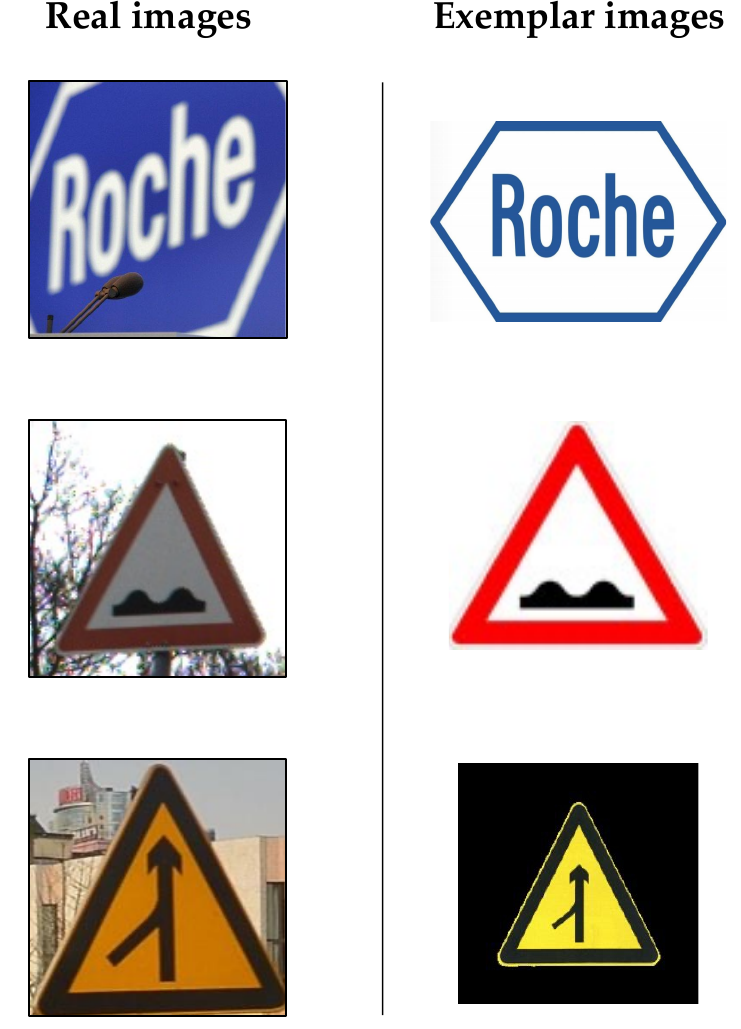}
    \caption{\textbf{Canonical Exemplars of real images.} Examples of real-world images of symbolic data such as traffic signs and brand logos (left column) along with their corresponding class-specific exemplar images (right column). These exemplars are readily available for such symbols and lie on a canonical domain devoid of perturbations of the real images.}
    \label{fig:exemplar}
\end{figure}

In order to overcome these drawbacks, we propose a different approach to detecting out-of-distribution samples in the few-shot setting, which utilizes \emph{reconstruction} as an auxiliary task to induce self-awareness in a few-shot classifier. This self-awareness would allow the classifier to better detect when it is presented with an out-of-distribution sample, as it would be able to recognize when it is unable to accurately reconstruct the input. However, naively applying reconstruction fails in the few-shot setting due to overfitting. Inspired by \cite{kim2019variational}, we propose using reconstruction of \textit{class-specific exemplars}, instead of self-reconstruction, to flag out-of-distribution samples. Such exemplars act as ideograms to effectively encode the semantic information of the class it belongs to. Consequently, they serve to anchor the representations of in-distribution classes when access to a large number of samples is restricted. Many real-world graphical symbols, such as traffic signs and brand logos, have well-defined exemplars that lie on a simpler or canonical domain. Some examples are shown in Fig. \ref{fig:exemplar}. However, exemplars may not be available for all datasets, and for such cases, we provide a simple scheme to estimate them from the few-shot data in the \emph{embedding space}, without any changes to the algorithmic formulation.


Building on this idea of reconstruction, we propose a new meta-learning strategy to tackle the few-shot open-set recognition task. In the meta-training phase, we use episodes sampled from a base set, with each episode simulating a token few-shot open-set task. Specifically, each episode is created by randomly selecting a set of classes and populating a support set with limited samples belonging to those classes. A query set is created in a similar fashion but contains samples from classes both seen in the support set and beyond (see Fig. \ref{fig:prob_setup}). These episodes are subsequently used to train our framework: Reconstructing Exemplar-based Few-shot Open-set ClaSsifer (\textbf{ReFOCS}). Given an episode, ReFOCS projects these samples to a low-dimensional embedding space to perform a metric-based classification over the support classes. Simultaneously, a variational model is used to reconstruct the class exemplars to compute the reconstruction error. The class scores, in addition to the reconstruction error, are then utilized to recognize the probability of the sample being out-of-distribution.

\subsubsection*{Main Contributions}
Our primary contributions are summarized below:
\begin{itemize}
    \item[$\bullet$] We develop a new reconstruction based meta-learning framework which utilizes class-specific exemplars to jointly perform few-shot classification and out-of-distribution detection.
    \item By utilizing reconstruction as an auxiliary task in our meta-learning setup we induce self-awareness in our learning framework, enabling it to self-reject out-of-distribution samples without relying on carefully tuned thresholds.
    \item[$\bullet$] We introduce a novel embedding modulation scheme to make the learned representations more robust and discriminative in the presence of scarce samples. A weighted strategy for prototype computation is also introduced for reducing intra-class bias.
    \item Our framework outperforms or achieves comparable performance to the current state-of-the-art on a wide array of FSOSR experiments, thereby establishing it as a new baseline for FSOSR.
\end{itemize}

\section{Related works}

\noindent \textbf{Few-shot learning.} Few-shot learning \cite{rohrbach2013transfer,raychaudhuri2020exploiting,guan2020zero} aims to learn representations that generalize well to novel classes with few examples. Meta-learning \cite{hospedales2020meta_survey} is the one of the most common approaches for addressing this problem and it is generally grouped into one of the following categories: \textit{gradient-based methods} \cite{maml,reptile} and \textit{metric learning methods} \cite{match_net,proto_net}. Typical gradient based methods, such as MAML \cite{maml} and Reptile \cite{reptile}, aim to learn a good representation that enables fast adaptation to a new task. On the other hand, metric-based techniques like Matching Networks \cite{match_net}, and Prototypical Networks \cite{proto_net} learn a task-specific kernel function to perform classification via a weighted nearest neighbor scheme. Our framework belongs to the latter class of methods, with the ability to work in the open-set scenario.\vspace{\baselineskip}

\noindent \textbf{Out-of-distribution detection.}
Also known as novelty or anomaly detection, the out-of-distribution task is commonly formulated as the detection of test samples that fall outside of the data distribution used in training. Hendrycks et. al. \cite{hendrycks17baseline} showed that softmax alone is not a good indicator of out-of-distribution probability but statistics drawn from softmax can be utilized to make assumptions of the "normalcy" of a test sample. Liang et. al. \cite{liang2020enhancing} re-calibrated output probabilities by applying temperature scaling and used virtual adversarial perturbations to the input to enhance the out-of-distribution capability of the model. Note that most out-of-distribution detection focuses on either detecting perturbed samples or out-of-dataset samples. In addition, these methodologies are not extendable to few shot settings. \vspace{\baselineskip}

\noindent \textbf{Open-set classification.} Open-set classification is slightly different from out-of-distribution detection in that it focuses on not only rejecting samples from unseen classes but also achieving proper classification of the seen categories. This is a much harder problem as opposed to just detecting perturbed/corrupted samples \cite{peeler}. Bendale et al. \cite{open_max} introduced OpenMax which uses extreme value statistics to re-calibrate the softmax scores of samples from unseen classes and reject out-of-distribution samples by thresholding their corresponding confidence score. G-OpenMax \cite{Gopenmax} combines OpenMax with a generative model to synthesize the distribution of all unseen classes. Recently, counterfactual image generation \cite{Neal_2018_ECCV} has been proposed to generate hard samples in an effort to build a more robust model. Kong et al. \cite{kong2021opengan} extends this idea by designing a generative adversarial network for the generation of out-of-distribution samples. Wang et al. \cite{wang2021energy} followed a similar strategy to OpenMax \cite{open_max} and introduced an energy-based classification loss for re-calibration of the softmax classification scores. Note that all these methods are in a fully supervised learning setting and are not directly applicable to few-shot learning. Liu et. al. \cite{peeler} first address open-set recognition under the few-shot setting. The proposed framework, titled PEELER, builds on top of prototypical networks \cite{proto_net} and uses episodic learning guided entropy maximization for regularizing the softmax classification scores. Unseen categories are detected by thresholding the maximum value of the softmax score of each out-of-distribution sample.
A more recent work SNATCHER \cite{snatcher} builds on top of PEELER \cite{peeler} and utilizes transformers \cite{vaswani2017attention} along with carefully crafted normalization schemes to improve feature level consistency of in-distribution samples. This enables SNATCHER to detect samples from open-set categories by simple distance thresholding in the embedding space. These methods are, however heavily dependent on the choice of the threshold. To eliminate this dependency, we propose to address few-shot open-set recognition by utilizing exemplar reconstruction as an auxiliary task in the meta-learning setup, such that model is imbibed with self-awareness regarding the openness of an input sample. Exemplars have been used in many computer vision tasks ranging from image recognition \cite{kim2019variational} to panoptic segmentation \cite{hwang2021exemplar}. Our work is the first to study there effectiveness on the FSOSR problem.


\begin{figure*}[ht]
    \centering
    \includegraphics[width=0.95\textwidth]{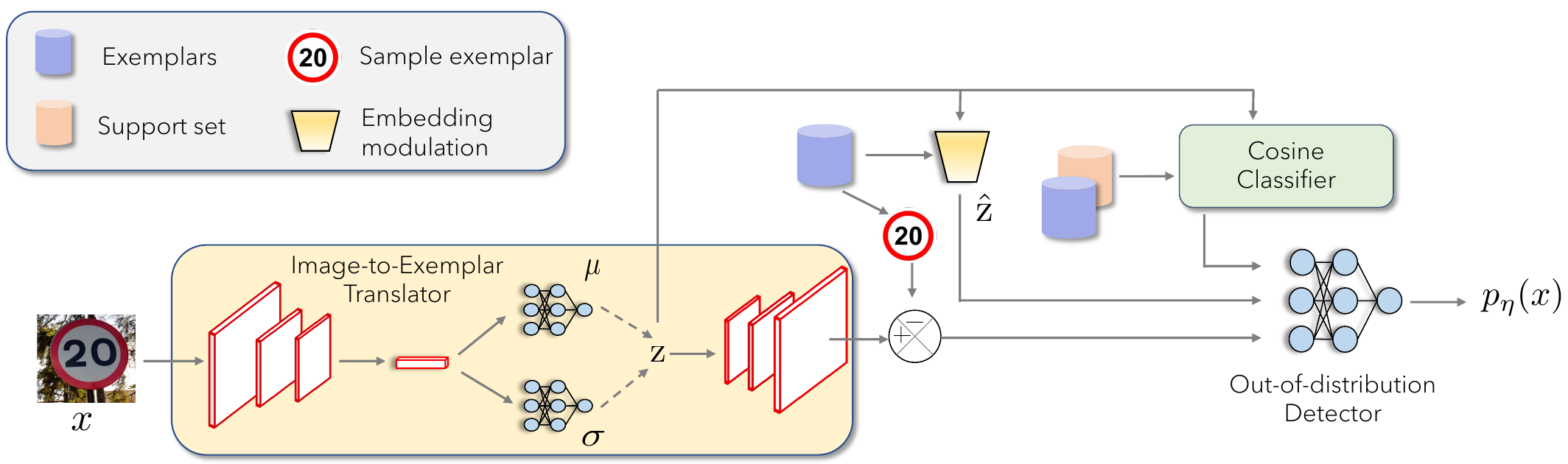}
    \caption{\textbf{Overview of framework.} Given a query sample $\mathbf{x}$, a latent representation $\mathbf{z}$ is derived by sampling from the variational posterior, the parameters of which are $\mu$ and $\sigma$. This embedding is further enhanced via a modulation process to get $\hat{\mathbf{z}}$. The latent embedding is used for classifying the sample into one of the few-shot classes. The decoder reconstructs the exemplar, $\mathbf{t}$ associated with the sample's class. The exemplar reconstruction error, the modulated embedding, and the classification scores themselves are fed to an MLP-based out-of-distribution detector to predict the probability $p_{\eta}$ of whether $\mathbf{x}$ is in/out-of-distribution with respect to the few-shot classes.}
    \vskip -0.1in
    \label{fig:framework}
\end{figure*}

\section{Methodology}
In this section, we present our framework for open-set few-shot classification. We first provide a  definition of the problem, followed by a brief overall methodology, and then present a detailed description of our framework.

\subsubsection*{Problem setting} 
Consider the standard few-shot learning setting, where we have access to a support set of labeled examples $\mathbb{S}=\left\{\mathrm{S}_1,\dots,\mathrm{S}_N\right\}$. Each $\mathrm{S}_c=\{\mathbf{x}_i\}_{i=1}^K$ denotes a set of $K$ examples belonging to the class $y_c$, for $N$-way $K$-shot recognition.  We make two changes to this setup. First, we assume the existence of class-specific exemplar images $\mathbf{t}_c$ for each $\mathbb{S}_c$; in case exemplars are not present, we estimate a class-specific exemplar for all the categories.  Second, the query set $\mathbb{Q}$ is comprised of both in-distribution and out-of-distribution samples w.r.t $\mathbb{S}$, i.e., $\mathbb{Q} = \mathbb{Q}_{in}\cup\mathbb{Q}_{out}$. In-distribution samples belong to classes seen in the support set while out-of-distribution samples belong to unseen classes. The goal is to detect the samples in $\mathbb{Q}_{out}$ as out-of-distribution, while correctly classifying the samples in $\mathbb{Q}_{in}$.

In order to develop a strong prior for few-shot learning, we utilize a base training set of labeled samples $\mathcal{B}=\{(\mathcal{X}_c,\mathcal{Y}_c)\}_{c=1}^M$ to meta-train our framework, where $M$ is the set of all classes in $\mathcal{B}$, $\mathcal{X}_c=\{\mathbf{x}_1,...,\mathbf{x}_{|c|} \}$ is the set of images in class $c$ and $\mathcal{Y}_c$ denotes the $c^{th}$ class label. If class-specific exemplars $\mathbf{t}_c$ are provided, we add them to $\mathcal{B}$ to obtain $\mathcal{\hat{B}}=\{(\mathcal{X}_c,\mathcal{Y}_c,\mathbf{t}_c)\}_{c=1}^M$, otherwise we first estimate $\mathbf{t}_c$ from $\mathcal{B}$ as shown in section 3.5 and then add them to $\mathcal{B}$ to get $\mathcal{\hat{B}}$ in a similar fashion. Like prior work \cite{proto_net,match_net} a set of training tasks or episodes $\{\mathcal{T}_i,\mathcal{T}_2,... \}$ are sampled from $\mathcal{\hat{B}}$ to conduct meta-training of the model. 
For an $N$ way $K$ shot problem, each episode $\mathcal{T}$ is constructed by first randomly sampling $N$ classes from $B$ and then constructing a support set $\mathbb{S}$ comprised of $K$ randomly chosen samples from each of the $N$ classes. Along with $\mathbb{S}$, a query set $\mathbb{Q}$ is constructed in a similar fashion. In order to simulate the presence of out-of-distribution samples, we follow the strategy outlined in \cite{peeler} and augment the query set with samples from classes \textit{absent} in the support.

\subsubsection*{Overall framework}
 A pictorial description of our framework is shown in Fig. \ref{fig:framework}. Given a sample $\mathbf{x}$, we first use variational inference to reconstruct the possible exemplar $\mathbf{t}$ associated with the ground-truth class of $\mathbf{x}$ and simultaneously obtain a latent representation of the same sample. This embedding is used to obtain a classification score, similar to \cite{chen2020new,match_net}, while the reconstructed exemplar is used as a proxy for the out-of-distribution detection task. Specifically, if $\mathbf{x} \in \mathbb{Q}_{out}$, we hypothesize that the reconstruction will fail for all the exemplars of the support classes. Based on this hypothesis, the latent representation, classification scores, and reconstruction errors (with respect to support exemplars) are subsequently fed into a binary classifier to predict the probability of the sample $\mathbf{x}$ being out-of-distribution. 
 
\subsection{Exemplar reconstruction from real images} \label{rec_mod}
Since the class-specific exemplars $\mathbf{t}$ form a compact representation of the real world images belonging to that class, we hypothesize that the ability to reconstruct any exemplar belonging to in-distribution classes correlates positively with the sample being an in-distribution one. Inspired by \cite{kim2019variational}, we use a Variational Auto-Encoder (VAE) \cite{Kingma2014} for the purpose of such reconstruction. The choice of VAE is motivated by its robustness to outliers and better generalization to unseen data, as shown in \cite{dai2018connections}. This is ideal for the image-to-exemplar translation task where the exemplars lie on a canonical domain devoid of the perturbations in real images. 

Given an input sample $\mathbf{x}$, its exemplar reconstruction of $\mathbf{t}$ is carried out by maximizing the variational lower bound of the likelihood $p({\mathbf{t}})$ \cite{kim2019variational} as follows,
\begin{equation}\label{elbo}
     \log p(\mathbf{t}) 
     \geq \mathbb{E}_{q_{\phi}(\mathbf{z}|\mathbf{x})} [\log   p_{\theta}(\mathbf{t}|\mathbf{z})] - D_{KL}[q_{\phi}(\mathbf{z}|\mathbf{x})||p(\mathbf{z})]
\end{equation}
where $D_{KL}[.]$ is the Kullback-Leibler (KL) divergence and $q_{\phi}(\mathbf{z}|{\mathbf{x}})$ denotes the variational distribution introduced to approximate the intractable posterior. Note that this is different from a vanilla VAE \cite{Kingma2014}, which is derived by maximizing the log-likelihood of the input data $\mathbf{x}$. A differentiable version of the lower bound is derived by assuming the latent variable $\mathbf{z}$ to be Gaussian in nature, sampled from the prior $q_{\phi}(\mathbf{z}|{\mathbf{x}})$. The empirical loss to be minimized is given as follows, 
\begin{equation}\label{recon_loss}
    \mathcal{L}_{VAE} = \frac{1}{K_e}\sum_{i=1}^{K_e}-\text{log}\ p_{\theta}({\mathbf{t}}_i|\mathbf{z}_i)+D_{KL}[q_{\phi}(\mathbf{z}|{\mathbf{x}})||p(\mathbf{z})]
\end{equation}
where $K_e$ is the number of samples in one episode, i.e., $K_e=|\mathbb{S}\cup\mathbb{Q}_{in}|$. Since $\mathbf{z} \sim q_{\phi}(\mathbf{z}|{\mathbf{x}})$ is non-differentiable, the re-parameterization trick is applied via the decoder network \cite{Kingma2014} such that $\mathbf{z} = \mathbf{\mu} + 
\mathbf{\sigma}\odot\mathbf{\epsilon}$, where $\mathbf{\epsilon} \sim \mathcal{N}(0,\textbf{I})$ and $\odot$ is the Hadamard product. 

The first term in Eq. \ref{recon_loss} is the reconstruction loss which affects the mapping of the real images to their class-specific exemplars, while the second term acts as a distribution regularization, enforcing the latent variable $\mathbf{z}$ to follow the chosen prior. Although binary cross entropy (BCE) is the common choice for the reconstruction loss in VAE, other losses such as $\ell_1$ or $\ell_2$ norm can also be used.

\subsection{Few-shot classification} \label{clf_mod}
The latent representation of a query sample $\mathbf{z}_q$, obtained from the encoder in the previous section, is used for computing the classification scores. We use the cosine metric to compute a relation score between the query sample and the support set $\mathbb{S}$. Specifically, the relation score is obtained by computing the cosine similarity between $\mathbf{z}_q$ and the set of prototypes or centroids, $\{\mathbf{\Omega}_{c}\}_{c=1}^{N}$ for each class $c \in \mathbb{S}$ ( Fig. \ref{fig:framework}). The classification score for $\mathbf{x}_q$ is obtained by performing a softmax operation on these relation scores. In contrast to prior works \cite{proto_net,peeler}, the class specific prototype $\mathbf{\Omega}_{c}$ is obtained by a weighted mean of the support samples instead of a simple mean. Note that these prototypes are different from the class specific exemplars $\mathbf{t}$.

\subsubsection*{Prototype computation} Given the support set, the prototypes for each class $c$ are calculated as follows,
\begin{equation}
    \mathbf{\Omega}_{c} = \sum\limits_{k=1}^{K}\omega_k \cdot \mathbf{z}_k^{c}
\end{equation}
$\mathbf{z}_k^{c}$ denotes the latent representation of $\mathbf{x}^k \in \mathbb{S}_c$, while
$\omega_k$ is the weight assigned to $\mathbf{z}_k^{c}$ based on how close it is from the embedding of the exemplar belonging to the $c^{th}$ class, $\mathbf{z}_t^{c}$,
\begin{equation}\label{proto_compute}
    \omega_k = \frac{e^{\cos(\mathbf{z}_k^{c},\mathbf{z}_t^{c})}}{\sum\limits_{k=1}^{K}e^{\cos(\mathbf{z}_k^{c},\mathbf{z}_t^{c})}}
\end{equation}
We use these weights in an effort to control the phenomena of \textit{intra-class bias} \cite{proto_rectify}, i.e.,  the difference between the true expected prototype and the Monte-Carlo estimated value. As explained earlier, each of the exemplars are an effective ideogram which provide a good abstraction of their respective classes. Thus the image-to-exemplar translation learned by the VAE leads to a feature space where the embedding of the real images cluster around that of their corresponding exemplars. This makes the exemplars good approximation of the true prototype and we leverage this via the weights $\omega_k$ to alleviate the intra-class bias.

\subsubsection*{Classification} After computing the prototypes, we predict the classification scores for the query sample $x_q$ as follows,
\begin{equation}
    p_{\phi}(y=c|\mathbf{x}_q) = \frac{e^{\tau \cdot cos(\mathbf{z}_q,\mathbf{\Omega}_c)}}{\sum\limits_{c^{'}}e^{\tau \cdot
    cos(\mathbf{z}_q,\mathbf{\Omega}_{c^{'}})}}
    \label{eq:classifier_prob}
\end{equation}
where $\tau$ is a learnable temperature parameter to scale the logits computed by cosine similarity  \cite{chen2020new,gidaris2018dynamic}. Learning proceeds by minimizing a cross-entropy loss over the in-distribution classes as follows:
\vspace{1pt}
\begin{equation}
    \mathcal{L}_{CE} =-\frac{1}{|\mathbb{Q}_{in}|}\sum_{i=1}^{|\mathbb{Q}_{in}|} \sum_{c=1}^{N} \mathbbm{1}{\{y_i=c\}}\log \ p_{\phi}(y=c|\mathbf{x}_{q,i})
    \label{eq:ce}
\end{equation}
where $y_i$ represents the true class of the query sample.

\subsection{Out of distribution detection}
Unlike prior work, we do not rely solely on the classification score to detect out-of-distribution query samples. Instead, ReFOCS flags query samples by leveraging the output of a multi-layer perceptron (MLP) classifier. This binary classifier takes into account three sources of information for scoring the openness of a query sample. These sources are (i) the class probability $\mathbf{p}_{\phi}$ as predicted in Eq. \ref{eq:classifier_prob}, (ii) a modulated version of the latent representation $\mathbf{z}$ (described below), and (iii) the set of reconstruction errors with respect to the support set exemplars, $\mathbf{D} =\left[||\hat{\mathbf{t}} -  \mathbf{t}_1||_F^2,...,||\hat{\mathbf{t}} -  \mathbf{t}_N||_F^2\right]$, where $\mathbf{D}$ indicates how far the reconstructed exemplar $\hat{\mathbf{t}}$ deviates from the actual exemplar $\mathbf{t} \in \mathbb{S}$. Intuitively, for out-of-distribution queries, all the entries of $\mathbf{D}$ will be very high, while for in-distribution samples, at least one of them will be very small. 

\subsubsection*{Embedding Modulation} At a fine-grained level, samples from many of the out-of-distribution classes can have very similar visual features to some of the in-distribution classes (Fig \ref{fig:softmax_issues}. This issue becomes more relevant for the few-shot setting since the model does not have access to large amounts of samples from the in-distribution classes for generalization. Therefore, given only a handful of samples from the in-distribution classes, it is of paramount importance to obtain an embedding that is discriminative enough to provide good segregation between in-distribution and out-of-distribution classes. This, in turn, will help the MLP in better detecting the out-of-distribution samples. While the latent embedding obtained from the VAE does have good discriminative properties, we introduce an additional modulation step that can enhance it even further. The enhanced embedding, $\mathbf{\hat{z}_q}$, is obtained by scaling the embedding of a query sample $\mathbf{z}_q$ with a scalar $\kappa > 0$ as shown  below,
\begin{equation}
    \mathbf{\hat{z}_q}=\frac{\mathbf{z}_q}{\kappa}, \ \  \text{where} \ \kappa = \min_{c \ \in \  \mathbb{S}}||\mathbf{z}_q - \mathbf{\Omega}_c||_1
\end{equation}
where $c \in \{1,...N\}$ represents the classes in the support and $\mathbf{\Omega}_c$ is the prototype corresponding to the $c^{th}$ class. $\kappa$ is a modulation factor measuring how close a query sample is to any of the in-distribution classes in the embedding space \cite{Savinov2019_EC}. out-of-distribution samples will tend to have higher values for $\kappa$ compared to in-distribution samples. Therefore this form of modulation will amplify the embeddings of in-distribution samples while scaling them down for out-of-distribution queries.


Therefore, the final input to the out-of-distribution detecting MLP is the concatenated vector $[\boldsymbol{\mathrm{p}}_{\phi}, \boldsymbol{\hat{\mathrm{z}}}_q, \mathbf{D}]$. This MLP classifier outputs a sigmoidal probability value $p_{\eta}$, indicating the openness of a query sample. Training proceeds by minimizing binary cross-entropy loss as follows,
 \begin{equation}
     \mathcal{L}_{BCE} = -\frac{1}{|\mathbb{Q}|}\sum\limits_{i=1}^{|\mathbb{Q}|}y_{\eta,i}\log \ p_{\eta,i}+(1-y_{\eta,i})\log (1-p_{\eta,i})
     \label{eq:bce}
 \end{equation}
where $y_{\eta}$ is equal to $0$ or $1$ depending on whether $x_q \in \mathbb{Q}_{in}$ or $x_q \in \mathbb{Q}_{out}$ respectively. 

\subsection{Training} \label{sec:training}
The parameters of the Encoder ($\phi$), Decoder ($\theta$) and the out-of-distribution detector ($\eta$) are jointly meta-trained by optimizing over the aggregate loss $\mathcal{L}$,
\vspace{1pt}
\begin{equation}\label{loss_func}
    \mathcal{L} = \lambda_1\mathcal{L}_{VAE}+\lambda_2\mathcal{L}_{CE}+\lambda_3\mathcal{L}_{BCE}
\end{equation}
where $\lambda_1$, $\lambda_2$ and $\lambda_3$ are hyper-parameters choices of which is discussed in Section \ref{Experiments}.

\subsection{Estimation of exemplars}
While it is easy to obtain well-defined exemplar images for images of graphical symbols, such class-specific exemplars may not always be provided for all kinds of natural images. We perform a simple exemplar estimation for categories that have missing exemplar images. First, we perform non-episodic training of the VAE encoder, $f_{\phi}$ on the entire base set $\mathcal{B}$. Second, the category-wise samples in $\mathcal{B}$ are passed through the pre-trained encoder to obtain corresponding feature representations, $f_{\phi}(\mathbf{x}_c)$ from the penultimate layer of the encoder. Finally, the exemplar image is defined via a nearest neighbor scheme in the feature space as follows,
\begin{equation}
    \mathbf{t}_{c} = \arg\min_{\mathbf{x} \in \mathcal{X}_c} ||f_{\phi}(\mathbf{x}) - \Psi_{c}||_2.
\end{equation}
Here, $\Psi_c = \frac{1}{|\mathcal{X}_c|}\sum\limits_{\mathbf{x} \in \mathcal{X}}{f_{\phi}(\mathbf{x})}$ is the centroid of the $c^{th}$ training class in the feature space.

For test episodes, we calculate the exemplar in a similar fashion by selecting the support sample closest to its centroid representation in the feature space.

\begin{table*}[ht]
\caption{\textbf{Episodic sampling strategy for each of the few-shot experiments.} For each episode $\mathcal{T}$, $\mathbf{K}_{\mathbb{Q}_{in}}^c$ denotes the number of in-distribution queries sampled from each support class $c \in \{1,..,N\}$ and the total number of in-distribution query samples is  \scalebox{0.75}{$\mathbf{K}_{\mathbb{Q}_{in}} = \sum\limits_{c=1}^N\mathbf{K}_{\mathbb{Q}_{in}}^c$}. The total number of OOD samples for each episode is denoted as $\mathbf{K}_{\mathbb{Q}_{out}}$. Hence, the total number of samples in each episode is $\mathbf{K}+\mathbf{K}_{\mathbb{Q}_{in}}+\mathbf{K}_{\mathbb{Q}_{out}}$ where, $\mathbf{K} = |\mathbb{S}|$.  The model is meta-trained with a total of $\mathbf{E}_{train}$ number of episodes and meta-tested on $\mathbf{E}_{test}$ number of episodes. }
\centering
\scalebox{0.9}{
\begin{tabular}{@{}lcccccccc@{}}
\toprule
\textsc{Datasets}  & \multicolumn{2}{c}{GTSRB$\rightarrow$GTSRB} & \multicolumn{2}{c}{GTSRB$\rightarrow$TT100K} & Belga$\rightarrow$Flickr32 & Belga$\rightarrow$Toplogos & \multicolumn{2}{c}{\textit{mini}ImageNet$\rightarrow$\textit{mini}ImageNet} \\ \midrule
\textsc{Experiment} & $5$-way $5$-shot     & $5$-way $1$-shot     & $5$-way $5$-shot      & $5$-way $1$-shot     & $5$-way $1$-shot           & $5$-way $1$-shot           & $5$-way $5$-shot                                      & $5$-way $1$-shot                                     \\ \midrule
$\mathbf{K}_{\mathbb{Q}_{in}}^c$     & 10                   & 10                   & 10                    & 10                   & 1                          & 1                          & 15                                                    & 15                                                   \\
$\mathbf{K}_{\mathbb{Q}_{out}}$      & 50                   & 50                   & 50                    & 50                   & 5                          & 5                          & 75                                                    & 75                                                   \\
$\mathbf{E}_{train}$                 & 20000                   & 20000                   & 20000                    & 20000                   & 50000                         & 50000                         & 50000                                                    & 50000                                                   \\
$\mathbf{E}_{test}$                  & 800                  & 800                  & 800                   & 800                  & 1700                       & 400                        & 600                                                   & 600                                                  \\ \bottomrule 
\end{tabular}
}

\label{tab:sampling}
\end{table*}
\section{Experiments}\label{Experiments}
In this section, we provide comprehensive experiments over several data sets to prove the efficacy of our framework, ReFOCS, for few-shot open-set classification. We compare the performance of ReFOCS with existing state-of-the-art methods that rely on thresholding approaches and softmax re-calibration for the detection of out-of-distribution samples. The overall results show that our method outperforms or achieves comparable performance to existing methods, without relying on the need for carefully tuned thresholds.


\subsubsection*{Datasets} 
We use six datasets to set up three different types of few-shot open-set recognition experiments: (i) \emph{traffic sign recognition}, for which we use the GTSRB \cite{gtsrb} and TT100K \cite{tt100k} datasets, (ii) \emph{brand logo recognition}, for which we use BelgaLogos \cite{belga1,belga2}, FlickrLogos-32 \cite{flickr} and TopLogo-10 \cite{toplogo}, and (iii) \emph{natural image classification} on the benchmark \textit{mini}ImageNet dataset \cite{match_net}. Different base training and meta-testing sets  are configured from these datasets to obtain five few-shot scenarios as shown in Tables \ref{tab:traffic}, \ref{tab:logo} and \ref{tab:miniimagenet}. Some of these scenarios involve cross-dataset experimentation, which is more challenging compared to using splits from the same dataset and better mimics real-world scenarios where training and test data can have significant domain shifts \cite{tseng2020cross}. It must be noted that all the traffic sign and logo datasets are provided with well-defined canonical exemplars for each of the classes. For these datasets the canonical exemplars can be directly used for the reconstruction task. To show the efficacy of our framework, we show results with both estimated and canonical exemplars for the traffic sign and logo datasets. The estimated exemplars are obtained following the strategy in section 3.5. Since \textit{mini}ImageNet is not provided with any well-defined exemplars, all of its class-specific exemplars are estimated. For the traffic sign and natural image datasets, we evaluate our model on both the $5$-way $5$-shot and the $5$-way $1$-shot tasks, while for the logo classification task, we show results only on the $5$-way $1$-shot scenario. This is due to BelgaLogos and Toplogos having multiple classes with less than $5$ samples. More details on these datasets and splits can be found in the supplementary material.


\subsubsection*{Implementation} 
The episodic sampling strategies for all the few-shot experiments are shown in Table \ref{tab:sampling}.
For the traffic sign and logo classification experiments, the VAE architecture is adapted from \cite{kim2019variational}. For the natural image classification task, we experiment with different resnet architectures \cite{resnet18} for the VAE encoder. In each case, the decoder is designed as an inverted version of the encoder architecture, for e.g., if Resnet18 \cite{resnet18} is used as the encoder, the decoder is designed as an inverted Resnet18. For all experiments, the MLP-based out-of-distribution detector is designed with two hidden layers, each containing $200$ and $100$ nodes, respectively. For a fair comparison, the encoder network is kept the same for all competing methods. The Adam optimizer \cite{kingma2017adam} is used for all the experiments. For all the traffic sign and brand logo recognition experiments, the values of the loss function hyperparameters, $\lambda_1,\lambda_2$ and $\lambda_3$, are set to $10^{-4},10$ and $10$ respectively. For the natural image classification task on \textit{mini}ImageNet $\lambda_1,\lambda_2$ and $\lambda_3$, are set to $10^{-4},1$ and $10$ respectively. The initial learning rate is set to $10^{-4}$ for both the traffic sign and brand logo experiments and to $10^{-3}$ for the natural image classification task. When using canonical exemplars for   the traffic sign and brand logo experiments the standard BCE criterion is used as the VAE reconstruction loss. On the other hand, for all experiments involving the estimated exemplars, we observed  using the $\ell_2$ norm as the reconstruction loss results in more improved performance. Additional implementation details are provided in the supplementary.


\begin{table*}[ht]
\centering
\tiny
\caption{\textbf{5-way 5-shot and 5-way 1-shot results of traffic sign recognition}. For both traffic sign datasets GTSRB$\rightarrow$GTSRB and GTSRB$\rightarrow$TT100K, $800$ test episodes were evaluated and the average performance is reported along with their $95\%$ confidence levels.}
\resizebox{0.9\textwidth}{!}{
\begin{tabular}{lcclllll}
\toprule
\multirow{3}{*}[-5pt]{\textsc{Model}} & \multirow{3}{*}[-5pt]{\textsc{Metric}} & \multirow{3}{*}[-5pt]{\textsc{Exemplar}}\\
& & & \multicolumn{2}{c}{\textsc{GTSRB}$\rightarrow$\textsc{GTSRB}} & \multicolumn{2}{c}{\textsc{GTSRB}$\rightarrow$\textsc{TT100K}} \\  
\cmidrule(l){4-7} \\
& & & \multicolumn{1}{c}{\textsc{Acc. (\%)}} & \multicolumn{1}{c}{\textsc{AUROC(\%)}} & \multicolumn{1}{c}{\textsc{Acc. (\%)}} & \multicolumn{1}{c}{\textsc{AUROC (\%)}} \\ 
\midrule
\multicolumn{4}{r}{5-way 5-shot} \\ \midrule
\textsc{ProtoNet \cite{proto_net}}    
& Euclidean  &   -                   
& 91.79 $\pm$ 0.44                      
& 70.74 $\pm$ 0.79           
& 80.51 $\pm$ 0.78    
& 62.10 $\pm$ 0.76  \\  

\textsc{Proto$+$OM \cite{open_max}}  
& Euclidean    &  -        
&  91.92 $\pm$ 0.43     
&  86.67 $\pm$ 0.48  
&  64.40 $\pm$ 0.78     
&  68.80 $\pm$ 0.66             \\
\textsc{PEELER \cite{peeler}}  
& Mahalanobis  &  -          
& 93.87 $\pm$ 0.37    
& 90.99 $\pm$ 0.44 
& 79.04 $\pm$ 0.79    
& 73.25 $\pm$ 0.81             \\

\textsc{ReFOCS}  
& Cosine     &  Estimated  
& 93.89 $\pm$ 0.48   
& 94.67 $\pm$ 0.25 
& 80.09 $\pm$ 0.79
& 79.48 $\pm$ 0.62    \\
\textsc{ReFOCS}  
& Cosine     &  Canonical
& \textbf{94.17 $\pm$ 0.38}   
& \textbf{94.83 $\pm$ 0.35}  
& \textbf{83.36 $\pm$ 0.76} 
& \textbf{85.25 $\pm$ 0.56}    \\ \midrule
\multicolumn{4}{r}{5-way 1-shot} \\ \midrule
\textsc{ProtoNet \cite{proto_net}} 
& Euclidean     &   -      
& 82.31 $\pm$ 0.75   
& 61.52 $\pm$ 0.90 
& 69.82 $\pm$ 0.87  
& 56.02 $\pm$ 0.83\\
\textsc{Proto$+$OM \cite{open_max}}  
& Euclidean     &   -      
& 82.46 $\pm$ 0.70  
& 81.43 $\pm$ 0.65 
& 55.10 $\pm$ 0.80    
& 67.79 $\pm$ 0.66           \\
\textsc{PEELER \cite{peeler}}
& Mahalanobis    &   -    
& 82.86 $\pm$ 0.77   
& 79.56 $\pm$ 0.85  
& \textbf{73.47 $\pm$ 0.88}   
& 69.68 $\pm$ 0.88       \\

\textsc{ReFOCS}        
& Cosine      &  Estimated
& 84.09 $\pm$ 0.71	
& \textbf{94.09 $\pm$ 0.21}     
& 70.13 $\pm$ 0.89	     
& 75.01 $\pm$ 0.76  \\ 
\textsc{ReFOCS}        
& Cosine      &  Canonical
& \textbf{86.21 $\pm$ 0.78}  
& 93.02 $\pm$ 0.45    
& 71.45 $\pm$ 0.87      
& \textbf{81.98 $\pm$ 0.59}   \\ 
\bottomrule
\end{tabular}
}

\label{tab:traffic}
\end{table*}
\begin{table*}[!ht]
\small
\centering
\caption{\textbf{5-way 1-shot results of brand logo recognition.} For \textbf{Belga$\rightarrow$Flickr32}, $1700$ test episodes were evaluated and for \textbf{Belga$\rightarrow$Toplogos}, $400$ test episodes were evaluated and their average closed-set Accuracy and open-set AUROC are reported with $95\%$ confidence intervals. }
\resizebox{0.85\textwidth}{!}{
\begin{tabular}{llcllll} 
\toprule
\multirow{2}{*}{\textsc{Model}} & \multirow{2}{*}{\textsc{Metric}}         &  \multirow{2}{*}{\textsc{Exemplar}} &  \multicolumn{2}{c}{\textsc{Belga}$\rightarrow$\textsc{Flickr32 }} & \multicolumn{2}{c}{\textsc{Belga}$\rightarrow$\textsc{Toplogos}~}       \\    
\cmidrule(l){4-7} \\
& & & \multicolumn{1}{c}{\textsc{Acc. (\%)}} & \multicolumn{1}{c}{\textsc{AUROC(\%)}} & \multicolumn{1}{c}{\textsc{Acc. (\%)}} & \multicolumn{1}{c}{\textsc{AUROC (\%)}} \\ 
\midrule
\textsc{ProtoNet} \cite{proto_net}                                 & Euclidean      & -                & 59.50 $\pm$ 0.99              & 58.69 $\pm$ 0.94     &       38.08~$\pm$~2.10              & 53.18~$\pm$~1.97                          \\
\textsc{Proto$+$OM} \cite{open_max}                                 & Euclidean    & -                  & 59.60 $\pm$ 1.01              & 61.00 $\pm$ 1.02     & 38.20 $\pm$ 1.95              & 56.40 $\pm$ 1.92                            \\
\textsc{PEELER}   \cite{peeler}                                  & Mahalanobis       & -              & 62.93 $\pm$ 1.03              & 66.40 $\pm$ 0.86       & 39.55~$\pm$~2.05              & 56.25~$\pm$~1.94~                         \\

\textsc{ReFOCS} & Cosine  & Estimated   
& 65.15 $\pm$ 1.02	& 69.08 $\pm$ 0.84 & 41.35 $\pm$ 2.07	& 57.36 $\pm$ 1.94 \\
\textsc{ReFOCS} & Cosine   & Canonical                        & \textbf{66.29 $\pm$ 1.02}     & \textbf{72.98 $\pm$ 0.83}        & \textbf{42.30~$\pm$~2.15}     & \textbf{58.39~$\pm$~1.97}         \\ 
\bottomrule
\end{tabular}
}

\vspace{-0.5em}
\label{tab:logo}
\end{table*}

\subsubsection*{Baselines} We compare ReFOCS against existing FSOSR methods such as PEELER \cite{peeler} and SNATCHER \cite{snatcher}. Both these methods are built on top of the Prototypical Networks (ProtoNet) \cite{proto_net} and rely on thresholding to reject out-of-distribution samples. Therefore, we also compare the performance of ReFOCS to that of ProtoNet, which although not designed for open-set recognition, provides an approximate lower bound on FSOSR performance. We also compare against OpenMax \cite{open_max}, which was originally proposed for open-set recognition in the fully supervised setting. OpenMax fits a Weibull distribution on the classification logits and utilizes its parameters to recalibrate the final softmax classification score. In order to adapt it for the few-shot setting, we implement OpenMax over the standard ProtoNet. We henceforth denote this baseline as Proto$+$OM. For both ProtoNet and Proto$+$OM, the out-of-distribution samples are rejected by using a the same thresholding principle as PEELER.

\subsubsection*{Evaluation Metrics} To quantify the closed-set classification performance, we compute the accuracy over in-distribution queries and utilize the Area Under the Receiver Operating Characteristic curve (AUROC) to quantify the model's performance in detecting the out-of-distribution samples. For all our evaluations, we split the query set equally among in-distribution and out-of-distribution samples.

\subsection{Traffic Sign Recognition}
\begin{figure}[!ht]
\centering
\captionsetup[subfigure]{justification=centering}
\subfloat[ ]{
	\label{subfig:exemplars}
	\includegraphics[width=0.47\textwidth]{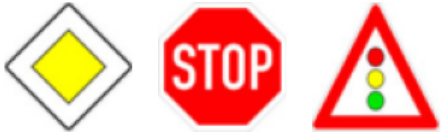} } 
\hfill
\subfloat[ ]{
	\label{subfig:in_query}
	\includegraphics[width=0.47\textwidth]{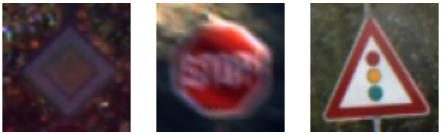} } 
\hfill
\subfloat[ ]{
	\label{subfig:out_query}
	\includegraphics[width=0.47\textwidth]{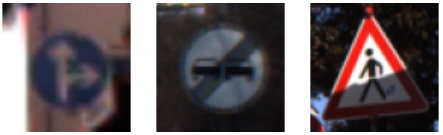} } 
\hfill
\subfloat[ ]{
	\label{subfig:inq_recon}
	\includegraphics[width=0.47\textwidth]{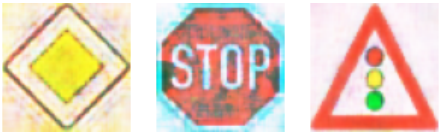} } 
	\hfill
\subfloat[ ]{
	\label{subfig:outq_recon}
	\includegraphics[width=0.47\textwidth]{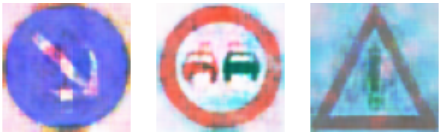} } 
	
\caption{\textbf{Sample exemplar reconstructions for GTSRB$\rightarrow$GTSRB.} Exemplar reconstructions of a few query samples from $1$ test episode. (a) Class-wise exemplars provided in the support set. (b) In-distribution query samples. (c) Out-of-distribution query samples. (d) Reconstructed exemplars from the in-distribution queries. (e) Reconstructed Exemplars from the out-of-distribution queries, as hypothesized when out-of-distribution samples are fed into ReFOCS it fails to reconstruct the in-distribution class-wise exemplars provided in the support.}
\label{fig:exemp_recon}
\end{figure}

The performance of ReFOCS on the two traffic sign recognition tasks, GTSRB$\rightarrow$GTSRB and GTSRB$\rightarrow$TT100K, is shown in Table \ref{tab:traffic}. GTSRB and TT100K have a total of $43$ and $36$ classes, respectively. For GTSRB$\rightarrow$GTSRB, $22$ classes are used for meta training, and the remaining $21$ are used for meta testing. For GTSRB$\rightarrow$TT100K, all classes of GTSRB are used for training, and testing is done on all the classes of TT100K. In both experiments, all images are resized to $64 \times 64$. As shown in Table \ref{tab:traffic}, ReFOCS outperforms all baselines using both estimated and canonical exemplars. Specifically for the task of detecting out-of-distribution samples, on average, ReFOCS achieves $\mathbf{7.4}$ percentage points higher AUROC compared to PEELER while using the estimated exemplars. On the other hand, using the well-defined canonical exemplars yields even greater performance gains up to nearly $\mathbf{10.4}$ percentage points compared to PEELER. The problem of just thresholding softmax classification scores to indicate the openness of samples can be clearly seen in the lower AUROC values of ProtoNet, Proto$+$OM, and PEELER. The issue is more pronounced for ProtoNet, since, unlike PEELER and Proto$+$OM, ProtoNet does not explicitly regularize the probability scores of out-of-distribution samples resulting in even more high-confidence false predictions. Utilizing our proposed prototype computation (Eq. \ref{proto_compute}) also results in higher classification accuracy on most of the FSOSR tasks. Since the canonical exemplars lie in a simpler domain they are devoid of many background and lighting perturbations seen in the estimated exemplars. As a result, the image-to-exemplar translation is much simpler for canonical exemplars compared to the estimated ones, which results in higher reconstruction errors for the latter, in turn affecting the out-of-distribution detection. This factor is even more amplified if there is a domain shift between the base training and meta-testing sets as in the case of GTSRB$\rightarrow$TT100K, which is why is comparison to GTSRB$\rightarrow$GTSRB, using the estimated exemplars over the canonical ones leads to a larger drop in AUROC for GTSRB$\rightarrow$TT100K. Nevertheless, the VAE-based reconstruction and our entire learning setup enable ReFOCs to better compensate for such domain shifts in comparison to existing FSOSR methods leading to significant increases in AUROC even while using the estimated exemplars.  Some sample exemplar reconstructions for the canonical exemplar case are provided in Fig \ref{fig:exemp_recon}.

\begin{table*}[t]
\centering
\caption{\textbf{5-way 5-shot and 5-way 1-shot results of natural image classification}. The best results for each backbone are shown in bold. The performance of all methods is averaged over $600$ test episodes and reported with $95\%$ confidence levels. When using the Resnet12 backbone, it is pre-trained for all competing methods using the same regimen as \cite{snatcher}.  $\dagger$ denotes our implementation using official code repository \textsuperscript{\ref{url}}.} 
\resizebox{0.85\textwidth}{!}{
\begin{tabular}{@{}lllcccc@{}}
\toprule 
                                \multirow{3}{*}{\textsc{Model}}
& \multirow{3}{*}{\textsc{Backbone}} & \multirow{3}{*}{\textsc{Metric}}& \multicolumn{4}{c}{\textit{mini}\textsc{ImageNet}$\rightarrow$\textit{mini}\textsc{ImageNet}} \\
                                & & & \multicolumn{2}{c}{$5$-way $5$-shot}                         & \multicolumn{2}{c}{$5$-way $1$-shot}   \\
                                \cmidrule(l){4-7} \\
                                
 & & & \multicolumn{1}{c}{\textsc{Acc.}(\%)} & \multicolumn{1}{c}{\textsc{AUROC}(\%)} & \multicolumn{1}{c}{\textsc{Acc.}(\%)} & \multicolumn{1}{c}{\textsc{AUROC}(\%)} \\ \midrule
\textsc{ProtoNet \cite{proto_net}}     
& Resnet18 & Euclidean           
& 78.11 $\pm$ 1.22  
& 58.67 $\pm$ 0.45  
& 55.18 $\pm$ 1.36    
& 56.51 $\pm$ 0.47\\  
\textsc{Proto$+$OM \cite{open_max}}  
& Resnet18 &Euclidean          
&  79.01 $\pm$ 1.21 
&  54.16 $\pm$ 0.47    
& 55.73 $\pm$ 1.36 
& 45.09 $\pm$ 0.39 \\

\textsc{PEELER} \cite{peeler} 
& Resnet18 &Mahalanobis    
& 75.08 $\pm$ 0.72        
& 69.85 $\pm$ 0.70   
& \textbf{58.31 $\pm$ 0.58}        
& 61.66 $\pm$ 0.62  \\
\textsc{PEELER\textsuperscript{$\dagger$} \cite{peeler}}  
& Resnet18 &Mahalanobis           
& 63.60 $\pm$ 0.67        
& 64.17 $\pm$ 0.64      
& 48.49 $\pm$ 0.81   
& 59.39 $\pm$ 0.77   \\
\textsc{ReFOCS} 
& Resnet18 &Cosine         
& \textbf{79.06 $\pm$ 1.17}   
& \textbf{69.91 $\pm$ 0.51}  
& \textbf{58.33 $\pm$ 0.84}   
& \textbf{69.24 $\pm$ 0.59} \\ 
\midrule

\textsc{ProtoNet \cite{proto_net}}     
& Resnet12 & Euclidean           
& 80.89 $\pm$ 1.18
& 59.14 $\pm$ 0.49 
& 65.02 $\pm$ 1.33   
& 52.47 $\pm$ 0.51  \\

\textsc{Proto$+$OM \cite{open_max}}  
& Resnet12 &Euclidean          
&  80.33 $\pm$ 1.22 
&  53.95 $\pm$ 0.51    
& 65.14 $\pm$ 1.39 
& 44.07 $\pm$ 0.36 \\

\textsc{PEELER}\textsuperscript{$\dagger$} \cite{peeler} 
& Resnet12 &Mahalanobis    
& 79.94 $\pm$ 0.63
& 73.58 $\pm$ 0.67
& 62.71 $\pm$ 0.79   
& 64.87 $\pm$ 0.80  \\

\textsc{Snatcher-F} \cite{snatcher} 
& Resnet12 &Euclidean    
& 82.02 
& \textbf{77.42} 
& 67.02 $\pm$ 0.85
& 68.27 $\pm$ 0.96  \\

\textsc{Snatcher-T} \cite{snatcher}  
& Resnet12 &Euclidean    
& 81.77
& 76.66
& 66.60$\pm$ 0.80 
& 70.17 $\pm$ 0.88  \\

\textsc{Snatcher-L} \cite{snatcher}  
& Resnet12 &Euclidean    
& \textbf{82.36}
& 76.15  
& \textbf{67.60 $\pm$ 0.83}  
& 69.40 $\pm$ 0.92  \\

\textsc{ReFOCS} 
& Resnet12 &Cosine         
& \textbf{82.61 $\pm$ 1.14}
& \textbf{77.31 $\pm$ 0.55} 
& 66.29 $\pm$ 0.91   
& \textbf{71.23 $\pm$ 0.62} \\
\bottomrule
\end{tabular}
}

\label{tab:miniimagenet}
\end{table*}
\begin{table*}[t]
\centering
\caption{\textbf{Ablation studies.} Recons. w/ AE refers to swapping the VAE with an Autoencoder (AE); Non-weighted prototype refers to the computation of the prototype as a simple centroid similar to ProtoNet \cite{proto_net}; No Modulation denotes turning off embedding modulation; No embedding/clf denote the removal of embedding/softmax scores from the input of the novelty module; ProtoC$+$ND refers to when we donot use exemplars for anything and consequently remove the reconstruction errors from the input of the novelty module.}
\resizebox{0.75\textwidth}{!}{
\begin{tabular}{@{}lcccc@{}}
\toprule
                                & \multicolumn{2}{c}{\textsc{GTSRB}$\rightarrow$\textsc{TT100K}}                 & \multicolumn{2}{c}{\textsc{Belga}$\rightarrow$\textsc{Flickr32}}              \\ 
\textsc{Experiment}                      & \multicolumn{2}{c}{$5$-way $5$-shot}                         & \multicolumn{2}{c}{$5$-way $1$-shot}                         \\ \midrule
                                & \multicolumn{1}{c}{\textsc{Acc.}(\%)} & \multicolumn{1}{c}{\textsc{AUROC}(\%)} & \multicolumn{1}{c}{\textsc{Acc.}(\%)} & \multicolumn{1}{c}{\textsc{AUROC}(\%)} \\ \midrule
Recons. w/ AE & 81.33 $\pm$ 0.80             & 77.08 $\pm$ 0.79              & 63.13 $\pm$ 1.05             & 71.86 $\pm$ 0.86              \\
Non-weighted Prototype         & 81.79 $\pm$ 0.77             & 84.93 $\pm$ 0.59              & 64.07 $\pm$ 1.02             & 72.11 $\pm$ 0.86              \\

No modulation         & 80.94 $\pm$ 0.79             & 78.55 $\pm$ 0.65              & 63.26 $\pm$ 1.03             & 56.69 $\pm$ 0.97              \\
ProtoC$+$ND     &    76.92 $\pm$ 0.83                         &       72.50 $\pm$ 0.61                     & 61.64 $\pm$ 1.03             & 50.01 $\pm$ 0.99              \\
No embedding     &    83.03 $\pm$ 0.75                         &       74.33 $\pm$ 0.66                     & 65.49 $\pm$ 1.00             & 67.79 $\pm$ 0.91              \\
No clf     &    82.71 $\pm$ 0.74                         &       82.85 $\pm$ 0.56                     & 64.51 $\pm$ 1.01             & 71.58 $\pm$ 0.87              \\
Full Model                           & \textbf{83.36 $\pm$ 0.76}    & \textbf{85.25 $\pm$ 0.56}     & \textbf{66.29 $\pm$ 1.02}    & \textbf{72.98 $\pm$ 0.83}     \\ \bottomrule
\end{tabular}
}

\vspace{-1em}
\label{tab:ablation}
\end{table*}

\subsection{Brand Logo Recognition}
The brand logo datasets consist of everyday images of commercial brand logos. In both few-shot tasks, Belga$\rightarrow$Flickr32 and Belga$\rightarrow$Toplogos, the Belga dataset consisting of $37$ classes, is used for meta-training. Flickr32 and Toplogos each have $32$ and $11$ classes, respectively. Since some of the classes have as low as $2$ samples, we restrict our experiments to the $5$-way $1$-shot scenario. In both cases, all images are resized to $64 \times 64$. Similar to the traffic sign recognition experiments, ReFOCS outperforms the existing baselines as evident from the results shown in  Table \ref{tab:logo}. Specifically, while using estimated exemplars, on average, ReFOCS achieves nearly $\mathbf{1.89}$ percentage points higher AUROC compared to PEELER and with the canonical exemplars, the average AUROC gains are $\mathbf{4.28}$ percentage points. The overall accuracy is also significantly higher compared to the  baselines. Specifically, when using the canonical exemplars we observe an average increase of $\mathbf{3.05}$ percentage points in accuracy over PEELER. These results establish the superiority of ReFOCS for FSOSR. Similar to the GTSRB$\rightarrow$TT100K, the domain shift between BelgaLogos and the other two brand logo datasets increases the difficulty of the task, especially when using the estimated exemplars. However, the overall results clearly show the superiority of ReFOCS over existing FSOSR methods in compensating for such domain shifts. 


\subsection{Natural Image Classification}
The \textit{mini}ImageNet dataset, introduced in \cite{match_net}, is a benchmark dataset used for evaluating the performance of models for few-shot natural image classification. This dataset has a total of $100$ classes, and as per the splits introduced in \cite{match_net}, $64$ classes are used for training, $16$ for validation, and the remaining $20$ for testing. The images are resized to the standard resolution of $84 \times 84$ \cite{match_net}. As mentioned previously, this dataset does not have any categorical exemplars, and therefore, for all its experiments, we use the estimated exemplars. As shown in Table \ref{tab:miniimagenet}, for this dataset, we show results with different encoder or backbone architectures. Specifically for the Resnet12 encoder, it is first pre-trained for all competing methods using the same setup as \cite{snatcher}. Following this, meta-training starts using the pre-trained weights of the encoder. From the overall results, we can observe that ReFOCS outperforms or achieves comparable performance to existing thresholding-based FSOSR methods. Since the images in \textit{mini}ImageNet have much more variations, methods like Proto$+$OM fail to achieve a decent AUROC score. This is because the Weibull distribution used by OpenMax \cite{open_max} under-fits to the handful of support samples for each of the novel test categories occurring in the meta-testing phase, which in turn renders it ineffective in recalibrating the softmax classification scores of the underlying Proto-Net. Since ReFOCS does not rely solely on the softmax scores for out-of-distribution detection it is able to achieve significantly higher AUROC scores. Using the pre-trained Resnet12 encoder for the meta-training phase boosts ReFOCS's performance even further. Pre-training also helps improve the performance of PEELER, although with the Resnet18 backbone, it fails to achieve its reported performance when the results are generated by using its official implementation  \footnote{\label{url}\url{https://github.com/BoLiu-SVCL/meta-open/}}.

In comparison to the variants of SNATCHER \cite{snatcher}, our framework is able to achieve comparable performance, even outperforming them in some cases. Since the authors of SNATCHER did not release their code we are able to compare with it for \textit{mini}ImageNet benchmark only, and its results shown in Table \ref{tab:miniimagenet} are taken directly from their paper \cite{snatcher}. Although SNATCHER is able to effectively use complex architectures like transformers \cite{vaswani2017attention} and carefully crafted normalization schemes to improve performance over PEELER, being a thresholding type FSOSR method, it is highly sensitive to the choice of an appropriate threshold. In contrast, ReFOCS eliminates the need for a hand-tuned threshold by utilizing the exemplar reconstruction setup to induce self-awareness in the model, making it more versatile than existing FSOSR baselines.

\begin{figure*}[ht]
\centering
\includegraphics[width=0.9\textwidth]{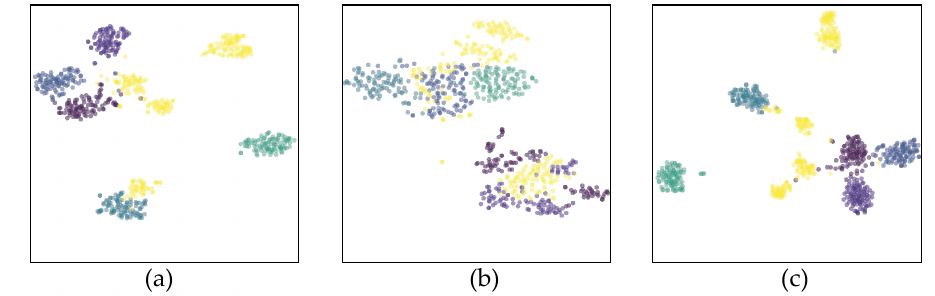}
\caption{\textbf{t-SNE visualization.} We project the latent space learned via (a) ProtoNet (b) PEELER (c) ReFOCS, on $5$ classes of GTSRB$\rightarrow$TT100K, on a 2D space using t-SNE. Out-of-distribution queries are in yellow}
\label{fig:tsne}
\end{figure*}

\subsection{Ablation Studies}
In this section, we use two cross-dataset scenarios to perform ablation studies of different components of our framework to understand their contribution toward the final performance.

\subsubsection*{Impact of using variational encoding.} The reconstruction module of ReFOCS is designed using a VAE \cite{kim2019variational} due to its better generalization ability \cite{dai2018connections}. We highlight this by replacing the VAE and experimenting with a standard auto-encoder (AE). As shown in Table \ref{tab:ablation}, both the classification and out-of-distribution detection performance drops significantly. This shows that the improved generalization capability of a VAE is necessary for handling FSOSR scenarios with domain shifts between the training and testing sets.

\subsubsection*{Impact of weighted prototype computation.}
By comparing the second and last row of Table \ref{tab:ablation} we can see that when our weighted prototype is replaced with a simple one as \cite{proto_net}, there is a considerable drop in the classification performance. This shows that utilizing the exemplar information in computing the weighted prototype results in more unbiased prototypical representations of each class, in turn facilitating improved metric-based classification of the in-distribution query samples.

\subsubsection*{Impact of embedding modulation.} The importance of the embedding modulation can be clearly seen in Table \ref{tab:ablation}. The modulated embedding results in more distinct feature representations with adequate segregation between the seen and unseen classes, thereby making it easier for the novelty module to flag unseen categories. Removing it results in a significant drop in AUROC, and on the other hand, its presence also amplifies the classification performance. This suggests that the improvement of the open class detection is complemented by improvement in seen class categorization.

\subsubsection*{Input to the Out-of-distribution detector.} We feed a concatenation of the modulated embedding, classification score, and $\ell_2$ reconstruction errors to the out-of-distribution detector. Empirically, the combination of all three gives the best results for out-of-distribution detection. We study the impact of removing the classification score or the embedding from the input. As seen from the results in Table \ref{tab:ablation}, in both cases, there is a drop in both the classification accuracy and the AUROC score. Additionally, we also examine the scenario when we do not use any exemplar for reconstruction and, consequently, feed in just the classification score and the raw embedding as input to the novelty module. We call this variant of ReFOCS, ProtoC$+$ND, and as seen from Table \ref{tab:ablation} the removal of the exemplar reconstruction errors results in the biggest drop in performance - with AUROC dropping to $50\%$ in one case (random prediction) - which again  consolidates the impact of the exemplars in out-of-distribution detection.

\subsubsection*{Choice of distance metric.} Computing the logits for classification requires a distance metric to measure the similarity between the prototypes and the sample in the latent space. We experiment with both the cosine and euclidean metrics and choose the cosine distance as it leads to more discriminative embeddings \cite{gidaris2018dynamic,proto_rectify} which is particularly important for segregating the open class samples from the support class ones. This is validated by the results shown in Table \ref{tab:metric}, where we can see that choosing the euclidean distance metric leads to a significant drop in AUROC.

\begin{table}[ht]
\centering
\caption{\textbf{Choice of metric.} Effect on performance as the similarity metric is varied. For \textit{mini}Imagenet the Resnet18 \cite{resnet18} backbone is used.}
\renewcommand{\arraystretch}{1.5}
\resizebox{0.48\textwidth}{!}{
\begin{tabular}{@{}lllcc@{}}
\toprule
\textsc{Dataset}  
& \textsc{Setting}
& \textsc{Metric}    
& \textsc{Acc.} 
& \textsc{AUROC} 
\\ \midrule
\multirow{2}{*}{GTSRB$\rightarrow$TT100K} 
& \multirow{2}{*}{5-shot}
& Cosine    
& \textbf{83.36$\pm$0.76}    
& \textbf{85.25$\pm$0.56}   
\\
&
& Euclidean 
& 78.77$\pm$0.74   
& 81.42$\pm$0.51  
\\
\midrule
\multirow{2}{*}{\textsc{Belga}$\rightarrow$\textsc{Flickr32}}  
& \multirow{2}{*}{1-shot}
& Cosine    
& \textbf{66.2$\pm$1.02}    
& \textbf{72.9$\pm$0.83}     
\\
&
& Euclidean 
& 48.73$\pm$0.99    
& 50.61$\pm$0.95     
\\ 
\midrule
\multirow{2}{*}{\textit{mini}\textsc{ImageNet}} 
& \multirow{2}{*}{5-shot}
& Cosine    
& \textbf{79.06$\pm$1.17}    
& \textbf{69.91$\pm$0.51}   
\\
&
& Euclidean 
& \textbf{79.09$\pm$1.15}    
& 52.83$\pm$0.48    
\\ 
\bottomrule
\end{tabular}}

\label{tab:metric}
\end{table}

\begin{table}[ht]
\centering
\tiny
\caption{Comparison of open recognition performance under different openness. Results are in terms of F1 score.}
\resizebox{0.48\textwidth}{!}{
\begin{tabular}{cccccc}
\midrule
&0\% &8.7\% &18.4\% &23.3\% &29.3\% \\
\midrule
& \multicolumn{5}{c}{\textsc{GTSRB}$\rightarrow$\textsc{TT100K}} \\
\midrule
PEELER& 59.11&	51.06&	48.91&	43.66&	40.28 \\
ReFOCS& \textbf{68.07}&\textbf{62.74}&\textbf{57.93}&\textbf{56.29}&\textbf{52.31} \\
\midrule
& \multicolumn{5}{c}{\textsc{Belga}$\rightarrow$\textsc{Flickr32}} \\
\midrule

PEELER& 48.91	&46.73	&41.22	&39.95	&37.14 \\
ReFOCS&\textbf{\textbf{53.65}}	&\textbf{50.52}	&\textbf{46.19}	&\textbf{43.78}	&\textbf{41.26} \\
\bottomrule

\end{tabular}}

\label{tab:openness}
\end{table}

\subsubsection*{Out-of-distribution performance with varying Openness}
Openness as defined in \cite{sun2020conditional} is shown below,
\begin{equation}
    openness = 1 - \sqrt{\frac{2N_{train}}{N_{test} + N_{target}}}
\end{equation}
where $N_{train}$ is the number of known classes seen during training, $N_{test}$ is the number of classes that will be observed during testing, and $N_{target}$ is the number of open classes to be recognized during testing \cite{sun2020conditional}. For an $N$-way $K$-shot problem, the support and the training set are the same, and therefore, $N_{train} = N_{test} = N$. $N_{target}$ refers to the number of open classes we sample $\mathbb{Q}_{out}$. We vary $N_{target}$ from $5$ to $15$ and observe the open-set recognition performance in terms of averaged F1 score \cite{sun2020conditional} for both ReFOCS and PEELER. As validated by Table \ref{tab:openness} ReFOCS consistently outperforms the thresholding-based PEELER on all the levels of openness.

\subsubsection*{Embedding Visualization.} In Fig. \ref{fig:tsne}, we compare the t-SNE \cite{tsne} visualization of the embedding spaces induced by all the competing methods. We can see from Fig. \ref{fig:tsne} that, in general, ReFOCS achieves more distinct class clusters in comparison to both PEELER \cite{peeler} and ProtoNet \cite{proto_net}, with adequate segregation between seen and unseen classes.


\section{Conclusion}
In this work, we present a novel strategy for addressing few-shot open-set recognition. We frame the few-shot open-set classification task as a meta-learning problem similar to \cite{peeler}, but unlike their strategy, we do not solely rely on thresholding softmax scores to indicate the openness of a sample. We argue that existing thresholding type FSOSR methods \cite{peeler,snatcher} rely heavily on the choice of a carefully tuned threshold to achieve good performance. Additionally, the proclivity of softmax to overfit to unseen classes makes it an unreliable choice as an open-set indicator, especially when there is a dearth of samples. Instead, we propose to use a reconstruction of exemplar images as a key signal to detect out-of-distribution samples. 
The learned embedding which is used to classify the sample is further modulated to ensure a proficient gap between the seen and unseen class clusters in the feature space. Finally, the modulated embedding, the softmax score, and the quality reconstructed exemplar are jointly utilized to cognize if the sample is in-distribution or out-of-distribution. 
The enhanced performance of our framework is verified empirically over a wide variety of few-shot tasks and the results establish it as the new state-of-the-art. In the future, we would like to extend this approach to more cross-domain few-shot tasks, including videos.
\vspace{-2em}
\section{Acknowledgement}
This work was partially supported by US National Science Foundation grant 2008020 and US Office of Naval Research grants N00014-19-1-2264 and N00014-18-1-2252.
\vspace{-1em}
\ifCLASSOPTIONcaptionsoff
  \newpage
\fi

\bibliographystyle{IEEEtran}
\bibliography{IEEEabrv,ref}

\vskip -2\baselineskip plus -1fil
\begin{IEEEbiography}[{\includegraphics[width=1in,clip,keepaspectratio]{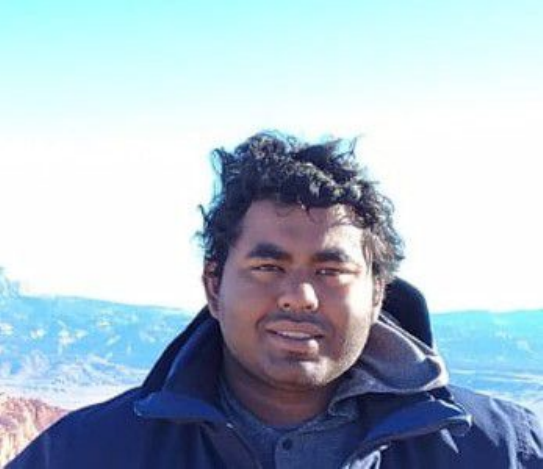}}]{Sayak Nag}
received his Bachelor’s degree in Instrumentation and Electronics Engineering engineering from Jadavpur University, Kolkata, India. Currently, he is pursuing a Ph.D. in the Department of Electrical and Computer Engineering at the University of California, Riverside. His broad research interests include computer vision and machine learning with a focus on few-shot learning, meta-learning, open-set recognition, and weakly-supervised learning.
\end{IEEEbiography}
\vskip -2\baselineskip plus -1fil
\begin{IEEEbiography}[{\includegraphics[width=1in,clip,keepaspectratio]{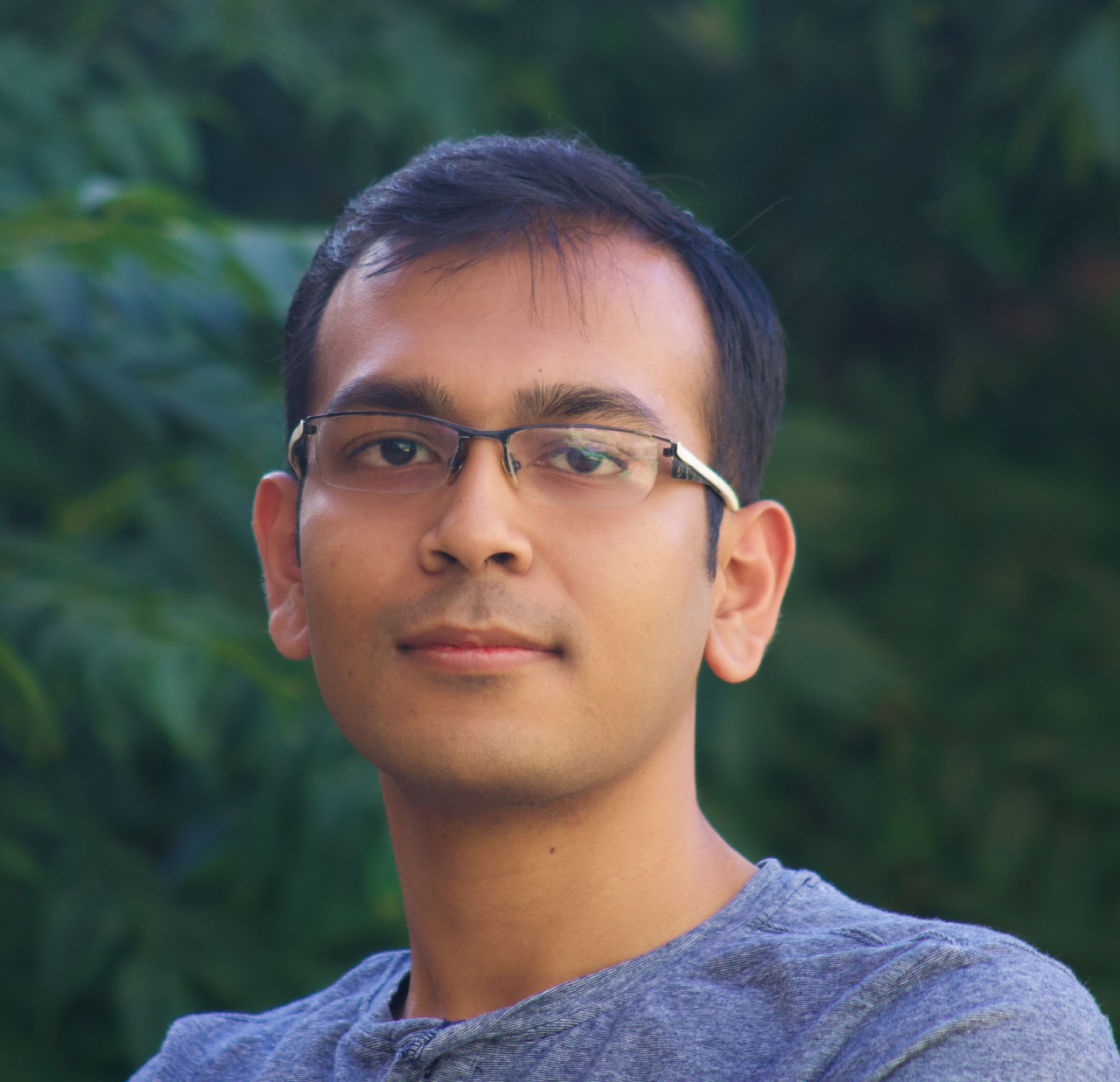}}]{Dripta S. Raychaudhuri} 
received his Ph.D. in Electrical and Computer Engineering from the University of California, Riverside, and his Bachelor’s degree in Electrical and Telecommunication engineering from Jadavpur University, Kolkata, India. He is currently an Applied Scientist at Amazon AWS, USA. His broad research interests include computer vision and machine learning with a focus on multi-task learning, domain adaptation, and imitation learning.
\end{IEEEbiography}
\vskip -2\baselineskip plus -1fil
\begin{IEEEbiography}
[{\includegraphics[width=1in,height=1.25in,clip,keepaspectratio]{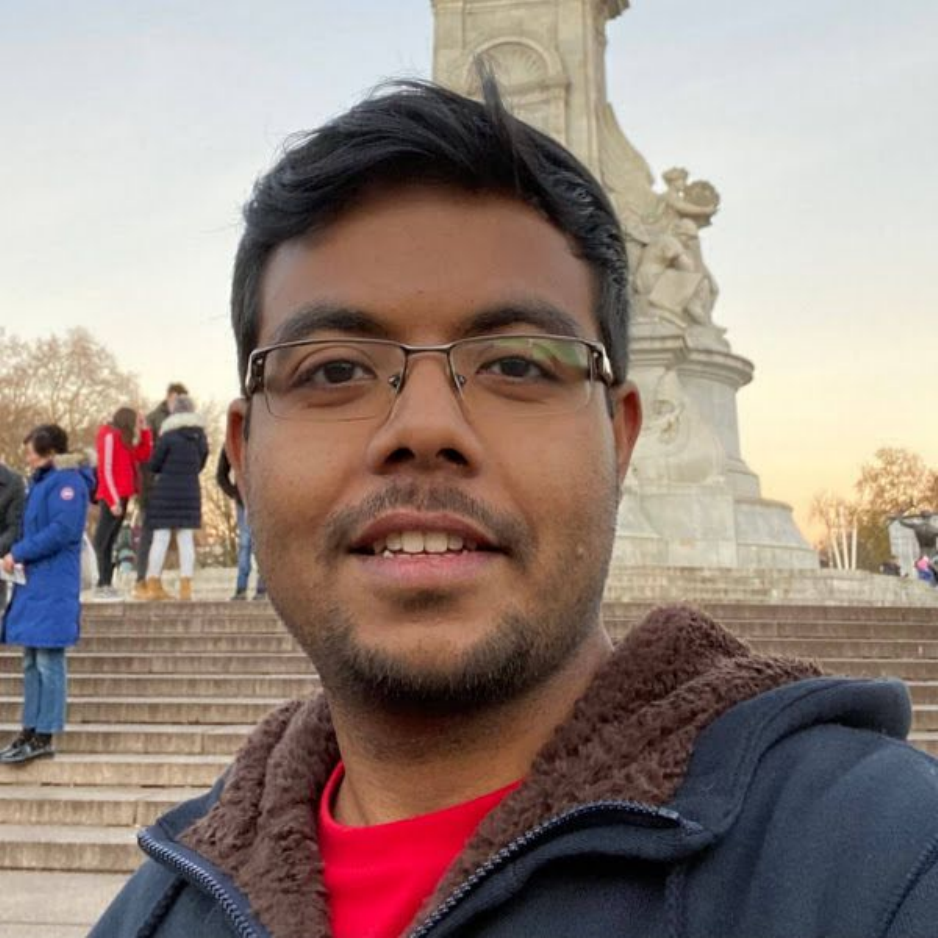}}]{Sujoy Paul} received his PhD in Electrical and Computer Engineering from the University of California, Riverside, and his Bachelor’s degree in Electronics and Telecommunication Engineering from Jadavpur University. 
He is currently a Research Scientist at Google Research, India. His broad research interest includes Computer Vision and Machine Learning, focusing on semantic segmentation, human action recognition, domain adaptation, weak supervision, active learning, reinforcement learning, and so on. 
\end{IEEEbiography}
\vskip -2\baselineskip plus -1fil
\begin{IEEEbiography}
[{\includegraphics[width=1in,height=1.25in,clip,keepaspectratio]{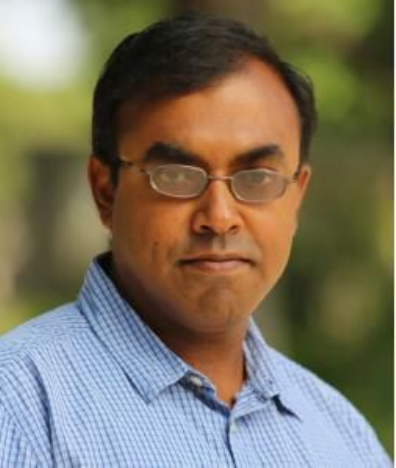}}]{Amit K. Roy-Chowdhury} 
received his PhD from the University of Maryland, College Park (UMCP) in 2002 and joined the University of California, Riverside (UCR) in 2004 where he is a Professor and Bourns Family Faculty Fellow of Electrical and Computer Engineering, Director of the Center for Robotics and Intelligent Systems, and Cooperating Faculty in the department of Computer Science and Engineering. He leads the Video Computing Group at UCR, working on foundational principles of computer vision, image processing, and statistical learning, with applications in cyber-physical, autonomous, and intelligent systems. He has published over 200 papers in peer-reviewed journals and conferences. He has also published two monographs on camera networks and wide-area tracking. He is on the editorial boards of major journals and program committees of the main conferences in his area. He is a Fellow of the IEEE and IAPR, received the Doctoral Dissertation Advising/Mentoring Award 2019 from UCR, and the ECE Distinguished Alumni Award from UMCP.
\end{IEEEbiography}
%

\end{document}


\title{Reconstruction guided Meta-learning for Few Shot Open Set Recognition (Supplementary material)}
    \maketitle
    \section{Datasets}
    In this section we go over the details of each dataset used in our experiments. A summary of the statistics of each dataset is shown in Table \ref{tab:data_size}.
    
    \subsection{Traffic Sign Datasets.}
    
    \noindent \textbf{GTSRB.} This dataset \cite{gtsrb} is one of the largest traffic-sign dataset. It is comprised of 43 classes broadly categorized under prohibitory, danger and mandatory categories. For GTSRB$\rightarrow$GTSRB, the training set contains a total of 39,209 images and the test set has a total of 12,630 images. These images have variations in illumination, shading as well as, resolution \cite{kim2019variational}.  
   
    \noindent \textbf{TT100K.} The Tsinghua-Tencent 100K (TT100K) \cite{tt100k} is a Chinese traffic sign detection dataset. This dataset has more than 200 categories, out of which we used the ones with valid annotation and a corresponding exemplar. Similar to \cite{kim2019variational} this amounts to a total of $36$ valid classes to work with.
    
    \subsection{Brand Logo Datasets}
    \noindent \textbf{BelgaLogos.} BelgaLogos \cite{belga1,belga2} is comprised of $10,000$ images of everyday brand logos commonly found in almost every aspect of daily life. The images have significant perturbations such as blurring, lighting and contrast variations as well as, occlusions. The dataset is also riddled with significant class imbalance as it contains classes with as little as $2$ samples making it suitable for only $1$-shot learning problems. Similar to \cite{kim2019variational} we collect $9,475$ images from BelgaLogos categorised among $37$ logo classes to construct our logo recognition dataset. The images of this dataset have a lot more variations compared to the traffic sign datasets, especially blurring and occlusion, which in-turn makes learning harder. 
    
    \noindent \textbf{FlickrLogos-32.} This dataset \cite{flickr} is comprised of a collection of images corresponding to the $32$ brand logos of Flickr.  We use the splits introduced in \cite{kim2019variational}, which is comprised of a total of $3,372$ cropped logo images. 
    
    \noindent \textbf{TopLogo-10.} This dataset \cite{toplogo} contains logo images from 10 brands related to popular cloth, shoes, and accessory brands. The logo images are obtained from their respective products, and similar to \cite{kim2019variational} the collected images were cropped from bounding box annotations and divided among $11$ classes where the ‘Adidas’ class is divided into ‘Adidas-logo’ and the ‘Adidas-text’. 
    
    \subsection{Natural Image Dataset}
    
    \noindent \textbf{\textit{mini}ImageNet.} Originally proposed by Vinyals et. al. \cite{match_net}, \textit{mini}ImageNet has become a benchmark dataset for few-shot classification \cite{proto_net,peeler}. It is derived from the larger ILSVRC-12 dataset \cite{imagenet} and is comprised of $100$ classes each having $600$ colored images, which are popularly resized to $84 \times 84$ \cite{proto_net,match_net}. We use the same splits as \cite{match_net}, which comprises a training set with $64$ classes, a validation set with $16$ classes, and a test set with $20$ classes.


\begin{table*}[h!]
\centering
\caption{\textbf{Number of samples and classes in each Dataset.}}
\begin{tabular}{@{}lcccccc@{}}
\toprule
                  & GTSRB    & TT100K   & BelgaLogos & Flickr-32 & TopLogos & \textit{mini}ImageNet \\ \midrule
Number of Samples & $51,839$ & $11,988$ & $9,585$    & $3,404$   & $848$    & $60,000$                               \\
Total Classes     & $43$     & $36$     & $37$       & $32$      & $11$     & $100$                                  \\ \bottomrule
\end{tabular}

\label{tab:data_size}
\end{table*}

    \section{Additional implementation details}
    \subsection{Setup details}
    For the traffic sign and brand logo recognition, the learning rate remains constant throughout the meta-training stage. For experiments on \textit{mini}ImageNet we drop the learning rate by a factor of 10 after every 20000 episodes.
    \subsection{Training details}
    For the traffic sign and brand logo classification, all modules of ReFOCS are meta-trained end-to-end. However, for \textit{mini}ImageNet training the entire framework end-to-end results in over-powering of the reconstruction module resulting in decreased classification accuracy. We suspect this problem is due to the difficult image-to-exemplar translation task for \textit{mini}ImageNet which is brought about by the fact that the estimated exemplars of \textit{mini}ImageNet tend to have much more variability compared to those of the traffic sign and brand logo datasets. In order to rectify this, we first meta-train the VAE encoder with just the cross-entropy classification loss (Eq. 6), following which the weights of the trained encoder are fixed and utilized to jointly train the decoder and the novelty detection module using a combination of the reconstruction loss (Eq. 2) and the novelty detection loss (Eq. 8). 

    For \textit{mini}ImageNet we also explore starting the meta-training phase by using the pre-trained encoder (also called backbone) when using Resnet-12. For both backbones the pre-trained encoder is directly used for the exemplar estimation process explained in section 3.5. For the Resnet12 case, the pre-training is done following \cite{snatcher} and during meta-training, the encoder's weights are initialized with those of the pre-trained encoder.

  \section{More Ablations}

\begin{table*}[ht]
\centering
\caption{Comparison of self-reconstruction and exemplar reconstruction guided meta-training for FSOSR. }
\resizebox{0.85\textwidth}{!}{
\begin{tabular}{@{}lcccc@{}}
\toprule
                                & \multicolumn{2}{c}{\textsc{GTSRB}$\rightarrow$\textsc{TT100K}}                 & \multicolumn{2}{c}{\textsc{Belga}$\rightarrow$\textsc{Flickr32}}              \\ 
\textsc{Experiment}                      & \multicolumn{2}{c}{$5$-way $5$-shot}                         & \multicolumn{2}{c}{$5$-way $1$-shot}                         \\ \midrule
                                & \multicolumn{1}{c}{\textsc{Acc.}(\%)} & \multicolumn{1}{c}{\textsc{AUROC}(\%)} & \multicolumn{1}{c}{\textsc{Acc.}(\%)} & \multicolumn{1}{c}{\textsc{AUROC}(\%)} \\ \midrule
Self-Reconstruction & 41.73 $\pm$ 0.86 & 49.92 $\pm$ 0.67 & 33.96 $\pm$ 1.13 & 48.55 $\pm$ 0.91  \\
Estimated Exemplar Reconstruction & 80.09 $\pm$ 0.79 & 79.48 $\pm$ 0.62 & 65.15 $\pm$ 1.02 & 69.08 $\pm$ 0.84 \\
Canonical Exemplar Reconstruction & 83.36 $\pm$ 0.76 & 85.25 $\pm$ 0.56 & 66.29 $\pm$ 1.02 & 72.98 $\pm$ 0.83 \\
\hline
\end{tabular}
}

\label{tab:exemplar_importance}
\end{table*}
\begin{table*}[ht]
\centering
\caption{Comparison of different distance metrics for exemplar reconstruction and modulation factor. For the current setup, the distance metrics proposed in Eq 10 and Eq 7 are used. }
\resizebox{0.75\textwidth}{!}{
\begin{tabular}{@{}lcccc@{}}
\toprule
                                & \multicolumn{2}{c}{\textsc{GTSRB}$\rightarrow$\textsc{TT100K}}                 & \multicolumn{2}{c}{\textsc{Belga}$\rightarrow$\textsc{Flickr32}}              \\ 
\textsc{Experiment}                      & \multicolumn{2}{c}{$5$-way $5$-shot}                         & \multicolumn{2}{c}{$5$-way $1$-shot}                         \\ \midrule
                                & \multicolumn{1}{c}{\textsc{Acc.}(\%)} & \multicolumn{1}{c}{\textsc{AUROC}(\%)} & \multicolumn{1}{c}{\textsc{Acc.}(\%)} & \multicolumn{1}{c}{\textsc{AUROC}(\%)} \\ \midrule
Case 1 & 79.84 $\pm$ 0.77 & \textbf{79.38 $\pm$ 0.63} & \textbf{65.08 $\pm$ 1.06} & \textbf{69.21 $\pm$ 0.84} \\
Case 2 & \textbf{79.95 $\pm$ 0.80} & 75.89 $\pm$ 0.62 & \textbf{65.02 $\pm$ 1.01} & 66.95 $\pm$ 0.88 \\
Case 3 & 79.43 $\pm$ 0.79 & 75.34 $\pm$ 0.68 & 64.91 $\pm$ 1.04 & 66.87 $\pm$ 0.85 \\
Current Setup & \textbf{80.09 $\pm$ 0.79} & \textbf{79.48 $\pm$ 0.62} & \textbf{65.15 $\pm$ 1.02} & \textbf{69.08 $\pm$ 0.84} \\
\hline
\end{tabular}
}

\label{tab:why_not_cosine}
\end{table*}
\subsection{Why exemplar reconstruction is important for FSOSR?}
As we have mentioned before, in few-shot learning, naively applying self-reconstruction (reconstructing the input query sample itself) will not enable the learning of an effective task-irrelevant representation. Therefore, it is necessary to anchor the samples of each class to a high-quality abstraction of that class, which we achieve using class-specific exemplars. To verify this, we compare self-reconstruction with exemplar reconstruction for the FSOSR tasks of GTSRB$\rightarrow$TT100K and Belga$\rightarrow$Flickr32. As observed from Table \ref{tab:exemplar_importance}, it can be seen that naively applying self-reconstruction leads to a catastrophic failure to generalize to FSOSR tasks as evident from the lower than random AUROC values as well as the significantly lower classification accuracies.

\subsection{Choice of metric for exemplar estimation and modulation factor $\kappa$}

For the nearest neighbor based exemplar estimation (Eq 10) and the modulation factor $\kappa$ (Eq 7) we choose the $\ell_2$ norm (euclidean distance) and the $\ell_1$ norm (manhattan distance), respectively. One might ask why a cosine similarity-based distance such as the following,
\begin{equation}
    Distance_{cosine}(\mathbf{x},\mathbf{y}) = 1 - \frac{\mathbf{x}^T \mathbf{y}}{||\mathbf{x} ||_2 || \mathbf{y}||_2}
\end{equation}
 was not used for these two.  This is because, for the exemplar estimation (Eq 10), we observe that the simple euclidean distance suffices to get the best exemplar. However, for the embedding modulation factor $\kappa$, it is necessary to use the $\ell_1$ norm. This is because if a cosine similarity-based distance is used for $\kappa$, it is constrained between 0 and 1, whereas utilizing the $\ell_1$ norm allows $\kappa$ to have $\geq 1$ values enabling greater scaling down of the embedding of out-of-distribution samples. To study this, we experiment by considering the following three cases,
\begin{itemize}
    \item  \textbf{Case 1}: In Eq 10 we use $Distance_{cosine}()$ as shown above and keep $\kappa$ the same as Eq 7.

    \item \textbf{Case 2}: For $\kappa$ we use $Distance_{cosine}()$ and keep the exemplar estimation same as Eq 10.

    \item \textbf{Case 3}: For both $\kappa$ and exemplar estimation we use $Distance_{cosine}()$.
\end{itemize}
We compare the results obtained from these three cases with the current setup of Eq 7 and Eq 10. From the observations tabulated in Table \ref{tab:why_not_cosine}, we can see that estimating the exemplars using the cosine distance, as opposed to the euclidean distance, does not significantly impact the Accuracy or AUROC. Therefore, choosing the euclidean distance suffices. On the other hand, if $Distance_{cosine}()$  is used to compute $\kappa$ (Case 2 and Case 3), we observe a significant drop in AUROC since the scaling of the query embedding becomes limited.

\begin{figure}[t]
\centering
\captionsetup[subfigure]{justification=centering}
\subfloat[ ]{
	\label{subfig:lmbda_1}
	\includegraphics[width=0.47\textwidth]{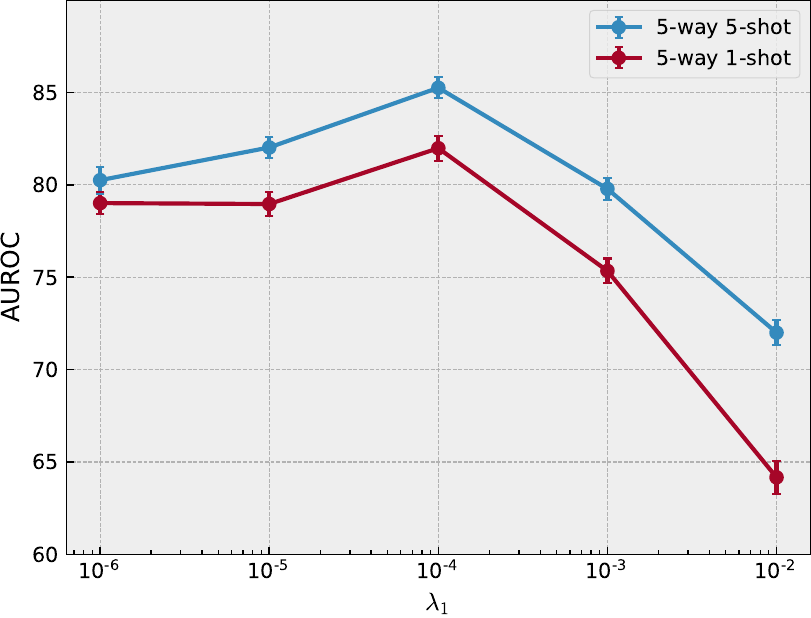} } 
\hfill
\subfloat[ ]{
	\label{subfig:lmbda_3}
	\includegraphics[width=0.47\textwidth]{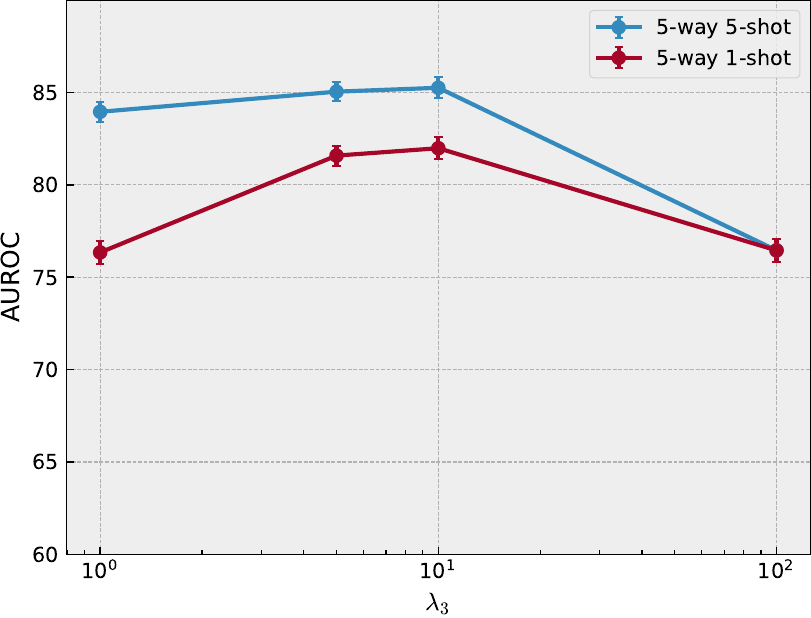} } 
\caption{\textbf{Hyperparameter Analysis}. (a) $\lambda_2$ \& $\lambda_3$ are fixed at $5$ and $10$,  changing only the reconstruction loss term. (b) $\lambda_1$ and $\lambda_2$ are fixed at $10^{-4}$ and $10$ and $\lambda_3$ is varied.}
\label{fig:hyper_params}
\end{figure}

  \subsection{Hyperparameter variation}
  Fig. \ref{fig:hyper_params} shows how the open-set AUROC is affected by the hyperparameters $\lambda_1$ or $\lambda_3$ for the GTSRB$\rightarrow$TT100K task. In both cases, the classification term $\lambda_2$ is fixed at $10$. In Fig. \ref{subfig:lmbda_1}, when $\lambda_1$ is very low, the open-set detection is hampered due to poor quality of the reconstruction and when it is increased beyond $10^{-4}$ reconstruction becomes the sole objective of the model, thus the open-set detection again degrades. From Fig. \ref{subfig:lmbda_3} we can see that for both the $5$-shot and $1$-shot cases increasing $\lambda_3$ causes the AUROC to improve the knee point of $\lambda_3=10$, after which it starts to degrade.

\section{More Qualitative Visualizations}

\begin{figure}[ht]
\centering
\vspace{-1em}
\captionsetup[subfigure]{justification=centering}
\subfloat[ ]{
	\label{subfig:exemplars}
	\includegraphics[width=0.5\textwidth]{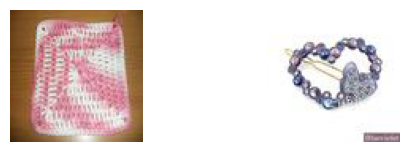} } 
\hfill
\subfloat[ ]{
	\label{subfig:in_query}
	\includegraphics[width=0.5\textwidth]{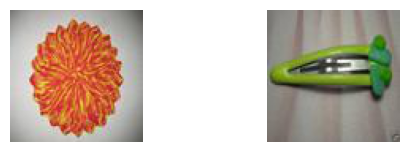} } 
\hfill
\subfloat[ ]{
	\label{subfig:out_query}
	\includegraphics[width=0.5\textwidth]{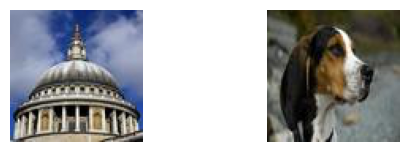} } 
\hfill
\subfloat[ ]{
	\label{subfig:inq_recon}
	\includegraphics[width=0.5\textwidth]{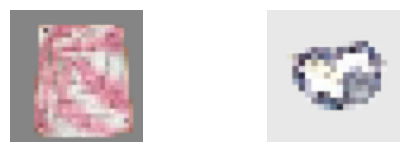} } 
	\hfill
\subfloat[ ]{
	\label{subfig:outq_recon}
	\includegraphics[width=0.5\textwidth]{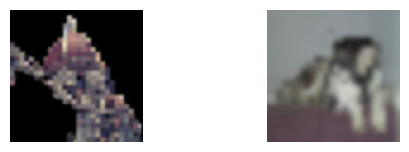} } 
	
\caption{\textbf{Sample exemplar reconstructions for \textit{mini}ImageNet$\rightarrow$\textit{mini}ImageNet.} Exemplar reconstructions of a few query samples from one test episode. (a) Class-wise exemplars are provided in the support set. (b) In-distribution query samples. (c) Out-of-distribution query samples. (d) Reconstructed exemplars from the in-distribution queries. (e) Reconstructed Exemplars from the out-of-distribution queries, as hypothesized when out-of-distribution samples are fed into ReFOCS it fails to reconstruct the in-distribution class-wise exemplars provided in the support.}
\label{fig:exemp_recon_mini}
\end{figure}

Some exemplar reconstructions for the \textit{mini}ImageNet $\rightarrow$ \textit{mini}ImageNet experiment are shown in Fig \ref{fig:exemp_recon_mini}. Since the exemplars are also natural images, the high variation in lighting and backgrounds makes the reconstruction much harder. However, ReFOCS performs the way it's intended to, and as such out-of-distribution samples fail to reconstruct the in-distribution exemplars.

\bibliographystyle{IEEEtran}
\bibliography{IEEEabrv,ref}